\documentclass[sn-mathphys,Numbered]{sn-jnl}

\usepackage{graphicx}%
\usepackage{multirow}%
\usepackage{amsmath,amssymb,amsfonts}%
\usepackage{amsthm}%
\usepackage{mathrsfs}%
\usepackage[title]{appendix}%
\usepackage{xcolor}%
\usepackage{textcomp}%
\usepackage{manyfoot}%
\usepackage{booktabs}%
\usepackage{algorithm}%
\usepackage{algorithmicx}%
\usepackage{algpseudocode}%
\usepackage{listings}%
\usepackage{subcaption}
\usepackage{orcidlink}

\theoremstyle{thmstyleone}%
\theoremstyle{thmstyletwo}%

\theoremstyle{thmstylethree}%

\begin{document}

\title[Article Title]{A novel automatic wind power prediction framework based on multi-time scale and temporal attention mechanisms}


\author[1]{Meiyu Jiang}\email{jiangmy21@lzu.edu.cn}
\author[2]{Jun Shen}\email{jshen@uow.edu.au}
\author[1]{Xuetao Jiang~\orcidlink{0000-0002-4818-6137}}\email{jiangxt21@lzu.edu.cn}
\author*[3]{Lihui Luo~\orcidlink{0000-0002-9471-7373}}\email{luolh@lzb.ac.cn}
\author[1]{Rui Zhou}\email{zr@lzu.edu.cn}
\author*[1]{Qingguo Zhou}\email{zhouqg@lzu.edu.cn}


\affil*[1]{\orgdiv{Department}, \orgname{Organization}, \orgaddress{\street{Street}, \city{City}, \postcode{100190}, \state{State}, \country{Country}}}

\affil[1]{\orgdiv{School of Information Science and Engineering}, \orgname{Lanzhou University}, \orgaddress{\city{Lanzhou}, \postcode{730000}, \state{Gansu}, \country{China}}}

\affil[2]{\orgdiv{School of Computing and Information Technology}, \orgname{University of Wollongong}, \orgaddress{\city{Wollongong}, \postcode{2522}, \state{NSW}, \country{Australia}}}

\affil[3]{\orgdiv{Gansu Provincial Industry Technology Center of Intelligent Equipment \& Big Data for Disaster Prevention, Northwest Institute of Eco-Environment and Resources}, \orgname{Chinese Academy of Sciences}, \orgaddress{\city{Lanzhou}, \postcode{730000}, \state{Gansu}, \country{China}}}


\abstract{Wind energy is a widely distributed, renewable, and environmentally friendly energy source that plays a crucial role in mitigating global warming and addressing energy shortages. Nevertheless, wind power generation is characterized by volatility, intermittence, and randomness, which hinder its ability to serve as a reliable power source for the grid. Accurate wind power forecasting is crucial for developing a new power system that heavily relies on renewable energy sources. However, traditional wind power forecasting systems primarily focus on ultra-short-term or short-term forecasts, limiting their ability to address the diverse adjustment requirements of the power system simultaneously. To overcome these challenges, We propose an automatic framework capable of forecasting wind power across multi-time scale. The framework based on the tree-structured Parzen estimator (TPE) and temporal fusion transformer (TFT) that can provide ultra-short-term, short-term and medium-term wind power forecasting power. Our approach employs the TFT for wind power forecasting and categorizes features based on their properties. Additionally, we introduce a generic algorithm to simultaneously fine-tune the hyperparameters of the decomposition method and model. We evaluate the performance of our framework by conducting ablation experiments using three commonly used decomposition algorithms and six state-of-the-art models for forecasting multi-time scale. The experimental results demonstrate that our proposed method considerably improves prediction accuracy on the public dataset Engie \url{https://opendata-renewables.engie.com}. Compared to the second-best state-of-the-art model, our approach exhibits a reduction of 31.75\% and 28.74\% in normalized mean absolute error (nMAE) for 24-hour forecasting, and 20.79\% and 16.93\% in nMAE for 48-hour forecasting, respectively.}

\keywords{Multi-time scale, Wind power forecasting, Transformer, Automatic framework}



\maketitle

\section{Introduction}\label{sec1}

Traditional energy sources pose significant environmental challenges and contribute to climate change. Burning coal, crude oil, and natural gas for electricity generation releases a significant amount of greenhouse gases, primarily carbon dioxide \cite{un_2022}. This harmful emissions contribute to the leading cause of global climate change. Furthermore, traditional energy sources have limited reserves and can be depleted if overexploited \cite{wang2019approaches}. Additionally, the prices of traditional energy sources tend to remain high and even increase over time. To address the energy crisis and its associated challenges, many countries are turning to renewable energy sources as a feasible alternative. Renewable energies, such as wind, solar, tidal, and biomass, offer several advantages. Firstly, these energy sources are abundantly available and widely distributed. Secondly, they are environmentally friendly since they do not emit greenhouse gases or contribute to climate change. Finally, renewable energy sources are often recyclable and sustainable.Among renewable energy sources, wind power has seen rapid development and increased efficiency. Converting wind energy into electricity has become more efficient and, consequently, less costly. Wind power has become a significant component of the global power generation infrastructure. The size and quantity of large wind farms are increasing rapidly \cite{luo2021local}. 

The increase in newly installed wind power capacity shows a continued commitment to transitioning to sustainable and renewable energy sources. According to the Global Wind Report 2022 \cite{gwec_2022}, the newly installed wind installations increased by 93.6 GW, slightly more than the 93 GW increase in 2020. Such developments in the wind energy sector are encouraging, as they contribute to the diversification of the energy mix, reduce greenhouse gas emissions, and mitigate climate change. The growing adoption of wind power around the world is a positive step towards a sustainable and cleaner energy future.

Wind power operates on the principle of converting wind energy into mechanical energy and subsequently converting it into electrical energy. Wind energy exhibits high levels of uncertainty, discontinuity, and frequent fluctuations. Specifically, wind turbine power generation is susceptible to instability and unpredictability, thus impacting power system scheduling and planning and frequently resulting in imbalances between power supply and demand. Addressing these issues necessitates the use of precise forecasts for forthcoming wind power generation, aiding in cost reduction and improved power generation efficiency. As a result, wind power technology is gaining significant significance within the renewable energy sector.

\begin{figure}[hbt!]
\centering
\includegraphics[width=1\textwidth]{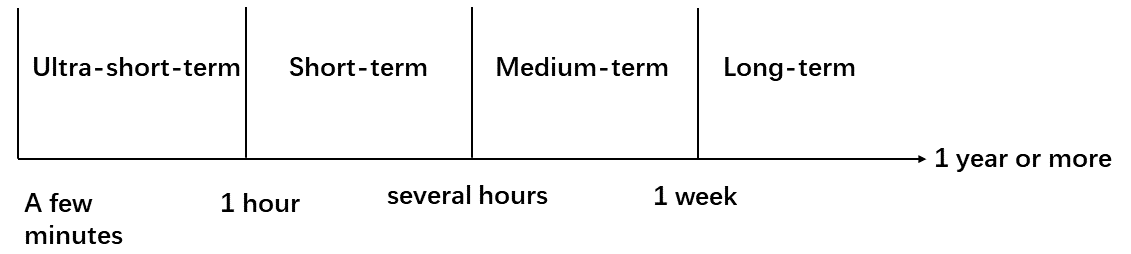}
\caption{Time-scale classification of wind power forecasting.}
\label{fig:scale}
\end{figure}

Figure \ref{fig:scale} depicts the classification of wind power forecasting models based on the time horizon, which includes ultra-short-term, short-term, medium-term, and long-term forecasts \cite{chang2014literature, aburiyana2017overview, devi2020hourly}. Time scale divisions vary among different literature sources, contributing to variations in specific wind power generation applications. The ultra-short-term forecasting model enables real-time monitoring of wind power generation. Short-term forecasting models assist power companies in formulating load dispatching plans, mitigating the effects of wind power grid integration throughout the entire system, and ensuring the secure operation of the electricity market. Medium-term forecasting models support activities including renewable energy trading, generation schedule optimization, and circuit maintenance. Long-term forecasting models aid in maintenance planning, including the optimization of wind farm locations and the development of annual generation plans. 

The majority of existing studies primarily focus on wind power forecasting for a single time horizon. Additionally, these studies often fall short in meeting the daily regulation requirements of power systems due to their concentration on ultra-short-term or short-term time horizons. The development of the power system primarily reliant on new energy sources necessitates the enhancement of the system's regulation capabilities. It is crucial to explore methods for achieving intelligent adjustment. To fulfill the ultra-short-cycle, short-cycle, and daily regulation requirements of the power system, this paper presents an automatic framework for wind power forecasting that offers forecasts for ultra-short-term, short-term, and medium-term intervals. It is worth noting that the accuracy of forecasts tends to diminish as the time horizon extends, introducing a particular challenge for medium-term wind power forecasting \cite{hanifi2020critical, chang2014literature, soman2010review}. 

\subsection{Contribution and paper organization}\label{subsec1}

This paper aims to tackle challenges in power systems by employing deep learning methods. Compared with conventional power sources, wind power can deliver significant amounts of clean energy. It is not reliable as a stable source of electricity. Directly integrating wind power generation into the power system poses challenges in maintaining power balance and restricting the flexibility of power dispatching. To address the issue of limited regulation capacity, it is imperative to develop an automatic wind power generation forecasting system that incorporates multi-time scale, ensuring the fulfillment of power supply and demand in the new power system.

Additionally, a substantial number of studies have neglected to validate their methods using publicly available datasets or selected limited testing dates \cite{wang2022deep, tian2022developing, duan2022novel}, consequently failing to capture the interconnections among various seasonal conditions \cite{xiong2022short, pei2022short, li2022wind}. Conversely, although the combination of decomposition algorithms and deep learning models can effectively capture nonlinear patterns in the data, the separate application of these methods hinders holistic tuning \cite{rayi2022adaptive, zhao2023hybrid, gao2023short, qiao2022wind}. Although multi-head attention mechanisms excel in natural language processing, they require supplementary mapping tasks to be suitable for time series forecasting. To address this issue, this paper presents a automatic framework that integrates attention mechanisms to optimize both the decomposition process and the model simultaneously. The primary contributions of this study are outlined as follows:
\begin{itemize}
  \item [(a)]
  This paper proposes an automatic framework for wind power forecasting that offers capabilities for ultra-short-term, short-term, and medium-term forecasting of wind power, and enables energy conversion between battery systems and grids.
  \item [(b)]
  This paper demonstrates the initial application of interpretable temporal fusion transformer (TFT) models to wind power generation. Existing deep learning models, such as temporal convolutional network, sequence to sequence, and transformer, treat all features uniformly. The TFT model categorizes them into static covariates, dynamic time-varying variables, and dynamic time-invariant variables as input features, enhancing the model's generalization performance.
  \item [(c)]
  This paper presents a tuning algorithm that utilizes the tree-structured parzen estimator (TPE) to iteratively optimize the hyperparameters of the decomposition algorithm and forecasting model. The TPE algorithm efficiently adjusts parameters, minimizing the likelihood of getting stuck in local optima and greatly influencing the discovery of the global optimum solution.
  \item [(d)]
  It demonstrates high accuracy on a public dataset throughout the entire year. Specifically, the proposed metnhod shows the lowest values of normalized mea absolute error (nMAE) and normalized root mean square error (nRMSE) achieved thus far.
  \end{itemize}

  The remaining subsections of this paper are organized as follows: Section 3 presents the fundamental theory and signal decomposition algorithms employed in this study. Section 4 details the proposed time series forecasting framework and the steps involved in the TPE-VMD-TFT method. Section 5 provides a demonstration of the framework's validity using the Engie wind dataset obtained from France's national power corporation. Finally, Section 6 concludes the paper and proposes future research directions.

\section{Literature review}\label{sec2}

According to modeling theory, wind power models can be classified into four groups: physical models, traditional statistical models, artificial intelligence-based models and hybrid models \cite{wang2021review}. The summary of selected studies on wind power forecasting is shown in Table \ref{tbl:sum}.

\begin{table}[h]
  \caption{Structure of the proposed framework.}
  \label{tbl:sum}
  \begin{tabular*}{\textwidth}{@{}cccclc@{}}
    \toprule
Classification &
  Author &
  \begin{tabular}[c]{@{}c@{}}Forecasted \\ areas\end{tabular} &
  \begin{tabular}[c]{@{}c@{}}Forecasting \\ horizon\end{tabular} &
  \multicolumn{1}{c}{\begin{tabular}[c]{@{}c@{}}Input \\ variables\end{tabular}} &
  \begin{tabular}[c]{@{}c@{}}Forecasting \\ methods\end{tabular} \\ \midrule
\begin{tabular}[c]{@{}c@{}}Physical \\ models\end{tabular} &
  \begin{tabular}[c]{@{}c@{}}Lazi$\acute{c}$ et al. \\ \cite{lazic2010wind}\end{tabular} &
  \begin{tabular}[c]{@{}c@{}}Eastern \\ Sweden\end{tabular} &
  48-h &
  \begin{tabular}[c]{@{}l@{}}wind speed, \\ wind shear,  \\  turbulence, \\ air density\end{tabular} &
  \begin{tabular}[c]{@{}c@{}}Regional \\ Eta model\end{tabular} \\
\multirow{2}{*}{\begin{tabular}[c]{@{}c@{}}Statistical \\ models\end{tabular}} &
  \begin{tabular}[c]{@{}c@{}}Yatiyana et al. \\ \cite{yatiyana2017wind}\end{tabular} &
  \begin{tabular}[c]{@{}c@{}}Western \\ Australia\end{tabular} &
  6-h &
  \begin{tabular}[c]{@{}l@{}}wind speed,\\ wind directions\end{tabular} &
  ARIMA \\
 &
  \begin{tabular}[c]{@{}c@{}}Firat et   al. \\ \cite{firat2010wind}\end{tabular} &
  Netherlands &
  \begin{tabular}[c]{@{}c@{}}24-h, \\ 48-h, \\ etc\end{tabular} &
  wind speed &
  \begin{tabular}[c]{@{}c@{}}ICA and \\ AR model\end{tabular} \\
\multirow{2}{*}{\begin{tabular}[c]{@{}c@{}}Artificial \\ intelligence \\ models\end{tabular}} &
  \begin{tabular}[c]{@{}c@{}}Bilal et al. \\ \cite{bilal2018wind}\end{tabular} &
  Senegal &
  1-min &
  \begin{tabular}[c]{@{}l@{}}wind speed, \\ solar radiation,\\ temperature, \\ humidity, \\ wind direction\end{tabular} &
  ANN \\
 &
  \begin{tabular}[c]{@{}c@{}}Jyothi   et al. \\\cite{jyothi2016very}\end{tabular} &
  North India &
  10-min &
  \begin{tabular}[c]{@{}l@{}}wind speed, \\ wind density,  \\ temperature, \\ wind direction\end{tabular} &
  AWNN \\
\multirow{2}{*}{\begin{tabular}[c]{@{}c@{}}Hybrid \\ models\end{tabular}} &
  \begin{tabular}[c]{@{}c@{}}Wang et al. \\ \cite{wang2020clustered}\end{tabular} &
  America &
  24-h &
  \begin{tabular}[c]{@{}l@{}}wind power, \\ wind speed, \\ temperature\end{tabular} &
  \begin{tabular}[c]{@{}c@{}}PSO-SVM-ARMA \\ model\end{tabular} \\
 &
  \begin{tabular}[c]{@{}c@{}}Shetty et al.\\ \cite{shetty2016optimized}\end{tabular} &
  India &
  1-h &
  \begin{tabular}[c]{@{}l@{}}wind speed, \\ wind direction,  \\ blade pitch angle, \\ density,\\ rotor speed\end{tabular} &
  \begin{tabular}[c]{@{}c@{}}RBF neural \\ network model\end{tabular} \\ \bottomrule
  \end{tabular*}%
  \end{table}

The physical approach does not require historical wind power generation data, but relies on the physical characteristics specific to the wind farm. Physical models estimate future wind power generation by predicting meteorological variables \cite{yan2019advanced}. The meteorological variables, typically wind speeds, can be derived from numerical weather prediction (NWP) or other meteorological factors such as temperature, humidity, and atmospheric pressure. Wind speed and wind power conversion can be obtained from wind power curves. The wind power curve is a graph that displays the power output of a wind turbine across various wind speeds. The curve is constructed by wind farm practitioners who measure wind speeds and the corresponding power output. Ackermann et al. \cite{ackermann2000wind} demonstrate that wind turbine power is proportional to the wind speed raised to the third power in wind power curves. This implies that a 10\% error in the predicted wind speed leads to a 30\% error in the predicted wind turbine power. Hence, accurate wind speed prediction is crucial for reliable wind power forecasting. Numerous studies have utilized physical model-based approaches for wind power forecasting. For instance, Lazi$\acute{c}$ et al. \cite{lazic2010wind} use the regional Eta model to predict wind speed and then predict wind power based on the wind power curve. The findings indicated that the Eta model could be employed as a reliable meteorological driver for wind power prediction.

Statistical models necessitate substantial amounts of data, including historical power values, historical weather data, and weather information, in order to train a model that accurately relates input and output power. Statistical models can be categorized into two types based on the parameters used: parametric models, liking time series models, and non-parametric models that rely on artificial intelligence techniques, such as artificial neural network (ANN) models and other data-driven models.

Time series models are advantageous for short-term or ultra-short-term wind power forecasting as they have the ability to capture fluctuating variations in output power. Autoregressive integrated moving average (ARIMA) and autoregressive (AR) models are commonly used in univariate time series forecasting. These models have a short computation time and are suitable for smooth time series, but may lead to inaccurate predictions when used with volatile input data. Several methods based on traditional statistical models are proposed to improve the accuracy of wind speed predictions. These methods typically use the wind power curve to convert wind speed to wind power. Yatiyana et al. \cite{yatiyana2017wind} develope a statistical method for predicting wind speed and direction using an ARIMA model. They validate the method using data from a specific site in western Australia. The experimental results indicate that the method achieve an error rate of less than 5\% for wind speed prediction, and less than 16\% for wind direction prediction. Firat et al. \cite{firat2010wind} introduce a statistical method for wind speed prediction that combines second-order blind identification (SOBI) with an autoregressive (AR) model. SOBI applies independent component analysis (ICA) to the input data, leveraging temporal structure to uncover hidden information or independent components. This approach yields more accurate predictions compared to direct wind speed prediction.

Physical models require detailed meteorological data and physical characteristics, placing higher demands on the dataset. Additionally, wind turbines are complex and modeling the wind-to-electricity conversion pattern is challenging. Traditional statistical models primarily handle linear relationships, which makes capturing the characteristics of non-linear wind power signal series challenging. With the development of artificial intelligence techniques, many ANN-based methods are proposed in the literature on wind power forecasting. Bilal et al. \cite{bilal2018wind} propose an ANN-based model for predicting wind turbine power. The input features are meteorological factors such as wind speed, wind direction, solar irradiance, temperature and humidity. The experimental results show that using meteorological factors as inputs to the ANN impacts the model's performance, with the most significant impact when wind direction and wind speed were used as feature inputs. Jyothi et al. \cite{jyothi2016very} use adaptive wavelet neural network (AWNN) for wind power generation. In addition to wind speed, wind direction, and ambient temperature, wind density is also considered as an input characteristic, with the authors using Morlet wavelets as the motor wavelets. The AWNN exhibited superior effectiveness compared to ANNs and the adaptive neuro-fuzzy inference system (ANFIS) when applied to wind power prediction problems.

Hybrid models combine different models to enhance the accuracy of wind power forecasting by leveraging the strengths of each model and taking into account various aspects of data fluctuations. Numerous studies utilize optimization algorithms to enhance the predictive performance of models. Wang et al. \cite{wang2020clustered} propose a cluster-based hybrid wind power forecasting model referred to as PSO-SVM-ARMA. Shetty et al. \cite{shetty2016optimized} propose the use of the radial basis function neural network (RBFNN) model for wind power prediction. They employ PSO to optimize the model's performance and leverage the extreme learning machine (ELM) to enhance the learning speed during training.

Existing medium-term wind power forecasting systems directly concat features such as historical wind power data, wind speed and wind direction, without considering the classification of input variables. The static characteristics of wind turbines, such as converter torque, generator converter speed, and pitch angle, are frequently overlooked. Bashir et al.'s study demonstrates that the properties of wind turbines have a profound influence on wind power generation \cite{bashir2022principle}. However, research on the categorization of input variables is currently insufficient. In contrast to prior studies, this paper introduces the use of the TFT model for the wind power generation prediction problem. It categorizes the input variables into static covariates, dynamic time-varying variables, and dynamic time-invariant variables. Additionally, it incorporates static features like converter torque to enhance the accuracy of wind power generation prediction.

In order to enhance the accuracy of wind power prediction models, numerous studies employ data pre-processing techniques to mitigate the instability of the raw input data. For instance, several research works propose the amalgamation of signal decomposition algorithms such as wavelet decomposition (WD), empirical mode decomposition (EMD), EEMD, and variational mode decomposition (VMD) with machine learning models. Among these techniques, VMD is widely recognized as the most effective time series decomposition approach \cite{zhao2023hybrid}. The VMD algorithm demonstrates strong adaptability as it can effectively decompose wind power generation series from different regions and climates. Rayi et al. \cite{rayi2022adaptive} employe the VMD algorithm to decompose wind data from three wind farms located in Sotavento, Spain, Wyoming, and California, USA. Similarly, Zhao et al. \cite{zhao2023hybrid} use the VMD algorithm to decompose wind speed series from a wind farm situated in Jiang County, Shanxi, China, consequently reducing data volatility.

\section{Methodology}\label{sec3}
\subsection{LSTM}\label{subsec3}

Hochreiter and Schmidhuber propose the LSTM model, a specific variant of a recurrent neural network (RNN) model \cite{hochreiter1997long}. This model not only captures contextual information and long-term dependencies encoded in time series data but also resolves the issue of gradient vanishing that commonly arises in RNN models. The LSTM comprises three distinct gates: the input gate, output gate, and forget gate, each with its unique function. The input gate regulates the flow of information into the memory cell, determining which portions of the current information stream are incorporated into the internal state of the cell. The cell state is then updates accordingly. Meanwhile, the forget gate decides which information should be discarded to accommodate new incoming data. Lastly, the output gate operates within the hidden layer, governing whether the information should be utilized as the output of the current LSTM. By selectively discarding irrelevant information and retaining salient details, the gating mechanism effectively addresses the problem of gradient dispersion encountered in traditional RNNs.

\subsection{XGBRegressor}\label{subsec3}

Unlike ANN, LSTM and CNN, which belong to strong learner, the eXtreme Gradient Boosting (XGBoost) algorithm proposed by Chen et al. is a Boosting algorithm with multiple learners \cite{chen2016xgboost}. XGBoost represents an advancement and commendable implementation of the gradient boosting decision tree (GBDT) algorithm by incorporating a regularization term into the GBDT objective function. This regularization term relates to factors like node partitioning difficulty and tree size, thereby enabling control over the tree size to diminish the risk of overfitting while expediting the algorithm's convergence. Moreover, the inclusion of second-order Taylor expansions facilitates accurate loss function computation and precise determination of the function's convergence direction. Being one of the few integrated learning algorithms that can rival strong learners, researchers have extensively investigated further aspects of the XGBoost algorithm, including techniques like missing value processing and feature importance analysis, thus making it highly valuable for addressing practical problems. Specifically, when XGBoost is applied to cope with regression tasks, it is commonly referred to as XGBRegressor.

\subsection{TFT}\label{subsec3}

\begin{figure}[hbt!]
  \centering
  \includegraphics[width=0.7\textwidth]{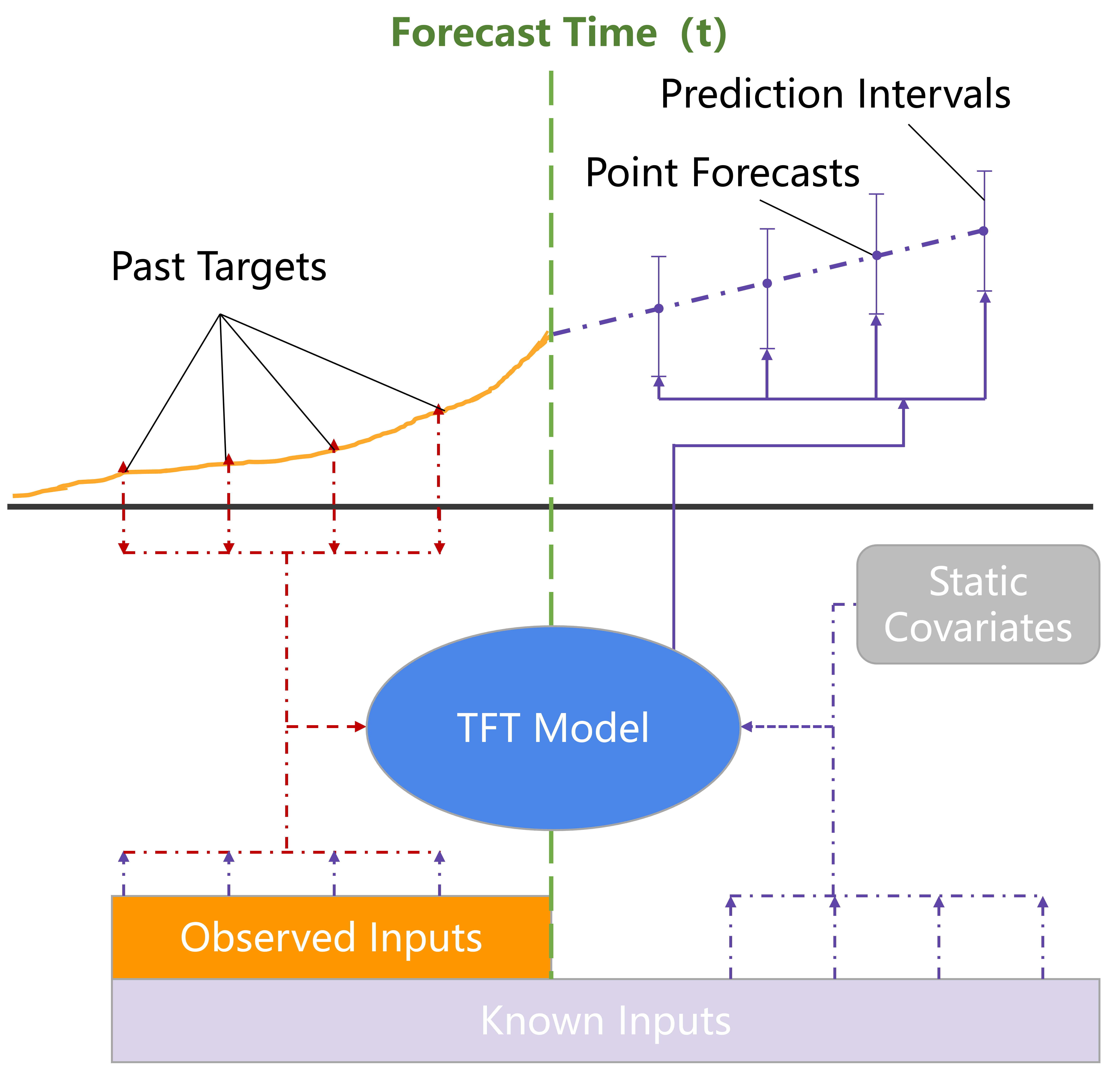}
  \caption{Process for multi-horizon time series forecasting}
  \label{fig:input}
\end{figure}

The TFT constitutes a deep neural network architecture that incorporates the attention mechanism. This architecture was originally proposed by Lim et al. and has gained significant traction in time series data prediction \cite{lim2021temporal}.  In comparison to other artificial intelligence-based models, the TFT model demonstrates remarkable performance improvements coupled with enhanced interpretability. As shown in Figure \ref{fig:input}, the TFT model classifies input features into distinct types, which include static covariates, future inputs that can be speculated upon, and time series data known from the past but not the future. Additionally, it captures interactions among the various types of input features during multi-horizon forecasting and accurately assesses the significance of these features for forecasting outcomes.

\begin{equation}
  \label{eq:step}
   \hat{y}_{i}(q,t, \tau) = f_{q}(\tau, y_{i,t-k:t}, o_{i, t-k:t}, k_{i, t-k:t+\tau}, s_{i})
\end{equation}

The multi-step prediction function, as presented in Equation \eqref{eq:step}, encompasses various variables for wind power forecasting. Notably, $i$ represents different wind farms, while $t$ denotes distinct points in time. Among the variables, $s_i$ corresponds to static variables that remain constant over time, while $y_{i,t}$ represents the target variable for prediction. The time-related dynamic variables, denoted as $x_{i,t} = \left[ o_{i,t}, k_{i,t} \right]$ can be classified into two categories. Specifically, $o_{i,t}$ signifies variables that fluctuate over time and remain unknown in advance, such as meteorological data (e.g., wind speed, wind direction, temperature, etc.). On the other hand, the variables indicated by $k_{i,t}$ vary over time yet can be deduced from known conditions, such as weeks, months, seasons and special holidays, etc. The forecasting model, denoted as $f_q$, along with the corresponding quantile function, estimates the value of quantile $q$ at time $t$ to forecast the future at step $\tau$. Distinctive from conventional regression models, the TFT employs the quantile loss metric instead of the traditional MSE loss. This enables the TFT model to adapt to multiple target regions in high-dimensional space. The quantile $q$ values may result in imbalances between positive and negative errors within the loss function. Selecting different $q$ are chosen, there is an imbalance between positive and negative errors in the loss function. Thus, the quantile loss aids in avoiding overfitting and underfitting of the model while supporting quantile regression. In the TFT model, the quantile group $Q$ is leveraged, and the final loss is determined by the weighted sum of all $q$ values within $Q$.

The TFT model comprises multiple purpose-specific modules that facilitate feature extraction from diverse types of input data. (1) Gating mechanisms serve to discard unnecessary information, simplifying the model structure while enhancing performance across various tasks. (2) Variable selection networks identify significant input features at each time step, mitigating the issue of overfitting in traditional DNNs. This, in turn, improves the model's capability to adapt to different samples and avoids predicting features irrelevant to the target. (3) Static covariate encoders play a role in moderating the temporal dynamics modeling by encoding static features. Recognizing that static information can be crucial for predicting the target, this module contributes to enhanced prediction outcomes. (4) Temporal processing empowers the model to capture both short and long-term time dependencies. This aspect encompasses two modules, namely the sequence-sequence layer and the multi-head attention layer. Together, they facilitate local processing and enable learning of long-term dependencies. (5) Prediction intervals, a result of quantile predictions, offer valuable insights into the output distribution, aiding in comprehensive understanding.

\subsubsection{Gating mechanisms}\label{subsubsec3}

In the field of time series forecasting, there are datasets of varying sizes and quality. To enhance the generalizability and adaptability of the TFT model to realistic and complex scenarios, the gated residual network (GRN) is employed to tackle the complexity of the input time series data. However, when dealing with small and noisy datasets, more complex models are unnecessary. In this case, GRN offers the flexibility to control the degree of non-linear transformation applied in order to improve model prediction. The GRN is defined as shown in Equation \eqref{eq:grn}, where its input consists of two types: the primary input data $a$ and the optional context vector $c$. In Equation \eqref{eq:eta2}, the activation function ELU is utilized, which is capable of taking negative values. Furthermore, ELU permits the unit activation means to be closer to 0 compared to other linear unsaturated activation functions like ReLU \cite{clevert2015fast}.

\begin{equation}
  \label{eq:grn}
  \operatorname{GRN}_{\omega}(\boldsymbol{a}, \boldsymbol{c})=\operatorname{LayerNorm}\left(\boldsymbol{a}+\operatorname{GLU}_{\omega}\left(\boldsymbol{\eta}_{1}\right)\right)
\end{equation}

\begin{equation}
  \label{eq:eta1}
  \boldsymbol{\eta}_{1}=\boldsymbol{W}_{1, \omega} \boldsymbol{\eta}_{2}+\boldsymbol{b}_{1, \omega}
\end{equation}

\begin{equation}
  \label{eq:eta2}
  \boldsymbol{\eta}_{2}=\operatorname{ELU}\left(\boldsymbol{W}_{2, \omega} \boldsymbol{a}+\boldsymbol{W}_{3, \omega} \boldsymbol{c}+\boldsymbol{b}_{2_{, \omega}}\right)
\end{equation}

The GLU plays a pivotal role in controlling the degree of non-linear transformation and enhancing the model's flexibility. It is mathematically defined as shown in Equation \eqref{eq:glu}. In this equation, the variable $\gamma$ represents the input, while $W$ and $b$ represent the weights and bias, respectively. Furthermore, the sigmoid activation function, denoted by $\sigma(.)$, is utilized in the GLU. The output of the sigmoid function falls within the range of 0 and 1, and it serves the purpose of feature selection. As an integral component of the GRN, the GLU is responsible for achieving a simple non-linear transformation when the input consists of small-scale data. In such cases, the GLU adjusts its output to remain close to zero.

\begin{equation}
  \label{eq:glu}
  \mathrm{GLU}_{\omega}(\boldsymbol{\gamma})=\sigma\left(\boldsymbol{W}_{4, \omega} \gamma + \boldsymbol{b}_{4, \omega} \right) \odot \left( {W}_{5, \omega} \gamma + \boldsymbol{b}_{5, \omega} \right)
\end{equation}

\subsubsection{Variable selection networks}\label{subsubsec3}

In cases where the model incorporates numerous input variables, determining the significance of the target vector in the prediction becomes challenging. To address this issue, the TFT model introduces Variable Selection Networks (VSNs) for selecting input variables. Instead of simply discarding features that have limited contributions to the prediction outcomes, the model assigns weights to the input variables. Higher weights are indicative of higher levels of importance. This approach not only alleviates the adverse effects of unimportant feature vectors but also enhances the model's performance.

\begin{equation}
  \label{eq:soft}
  v_{\chi_{t}}=\operatorname{softmax}\left(G R N_{v_{\chi}}\left(\Xi_{t}, c_{s}\right)\right)
\end{equation}

\begin{equation}
  \label{eq:sum}
  \widetilde{\xi}_{t}=\sum_{i=1}^{m_{x}} v_{\chi_{t}}^{(i)} \widetilde{\xi}_{t}^{(i)}
\end{equation}

\begin{equation}
  \label{eq:grnn}
  \tilde{\xi}_{t}^{(i)}=\operatorname{GRN}_{\tilde{\xi} (i)}\left(\xi_{t}^{(i)}\right)
\end{equation}

VSNs employ the GRN individually on each feature before concatenating them as input to the GRN once again. Additionally, softmax is utilized to calculate feature weights, which are then assigned to the respective input variables. The method for obtaining these weights is depicted in Equation \eqref{eq:soft}. The combination method for weighting the features is illustrated in Equation \eqref{eq:sum}, where $v_{\chi_{t}}^{(i)}$ represents the weight of feature selection, and $\tilde{\xi}_{t}^{(i)}$ signifies the feature after non-linear processing by the GRN. The formula for computing $\tilde{\xi}_{t}^{(i)}$ is described in Equation \eqref{eq:grnn}.

\subsubsection{Static covariate encoders}\label{subsubsec3}

The TFT model generates four output variables, namely $cs$, $cc$, $ch$ and $ce$ utilizing four distinct GRNs. (1) temporal variable selection ($cs$) as the input to Variable selection networks. (2) local processing of temporal features ($cc$, $ch$), using the LSTM as the initialization state. (3) enriching temporal features with static information ($ce$) and input into the Static Enrichment.

\subsubsection{Interpretable multi-head attention}\label{subsubsec3}

The interpretable multi-head attention module of the TFT model is built upon the transformer model's multi-head attention mechanism, as illustrated in Equations \eqref{eq:in} and \eqref{eq:h}. In the original transformer model, the multi-head attention mechanism creates multiple subspaces by partitioning the model into distinct heads. Each head learns different weights, enabling the model to capture diverse aspects of the features. Although aggregating these features yields improved predictions, interpreting them proves challenging.  To enhance interpretability, the TFT model introduces a modification wherein the weight-sharing scheme is applied to the $v$ matrix of each head, while the $Q$ and $K$ matrices retain their individual weights as before.
\begin{equation}
  \label{eq:in}
  InterpretableMultiHead  (\mathrm{Q}, \mathrm{K}, \mathrm{V})=\tilde{H} W_{H}
\end{equation}

\begin{equation}
  \label{eq:h}
  \begin{split}
  \begin{aligned}
  \tilde{H} & =\tilde{A}(Q, K) V W_{V} \\
  & =\left\{\frac{1}{m_{H}} \sum_{h=1}^{m_{H}} A\left(Q W_{Q}^{(h)}, K W_{K}^{(h)}\right)\right\} V W_{V} \\
  & =\left\{\frac{1}{m_{H}} \sum_{h=1}^{m_{H}} \text { Attention }\left(Q W_{Q}^{(h)}, K W_{K}^{(h)}, V W_{V}\right)\right\}
  \end{aligned}
  \end{split}
\end{equation}

\subsubsection{Signal decomposition algorithms}\label{subsubsec3}

Wu and Huang propose the EEMD algorithm as an improvement to the EMD method \cite{wu2009ensemble}. EEMD leverages the characteristic of white noise with a zero mean, enhancing the determination of extreme points in the signal by injecting Gaussian white noise multi-time during the EMD decomposition process. On the other hand, the VMD algorithm utilizes Wiener filtering, Hilbert transform, and frequency mixing for signal processing \cite{dragomiretskiy2013variational}.  Unlike the EMD algorithm, VMD is grounded in solid mathematical principles, allowing for accurate and efficient signal separation. To effectively extract multiple seasonal cycles from a time series, K. Bandara et al. introduce the Multiple Seasonal-Trend Decomposition based on Loess (MSTL) algorithm \cite{bandara2021mstl}. MSTL effectively decomposes the time series into multiple seasonal components, trend, and residuals.

\section{Proposed framework}\label{sec4}
\subsection{Rationale of the proposed framework}\label{subsec4}
The proposed framework incorporates the TPE-VMD-TFT method, which enables automatic wind power forecasting across various time scales. The TPE-VMD-TFT method uses VMD to decompose the wind generation series into subseries, TPE optimizes the key parameters of the decomposition algorithm and TFT model. To enhance the accuracy of wind power forecasting, an improved TFT model is employed. This model segregates historical wind power data, meteorological data, and static parameters of wind turbines into distinct types, which are then inputted into the model for training purposes. The TPE-VMD-TFT method leverages the TPE, VMD, and TFT models to facilitate an automatic prediction process, encompassing data preprocessing, model prediction, automatic hyperparameter optimization, and ultimately obtaining the optimal solution. With the detailed rationale of this framework being elaborated upon in the subsequent sections.

\subsection{Overview of the proposed framework}\label{subsec4}

\begin{figure}[hbt!]
  \centering
  \includegraphics[width=1\textwidth]{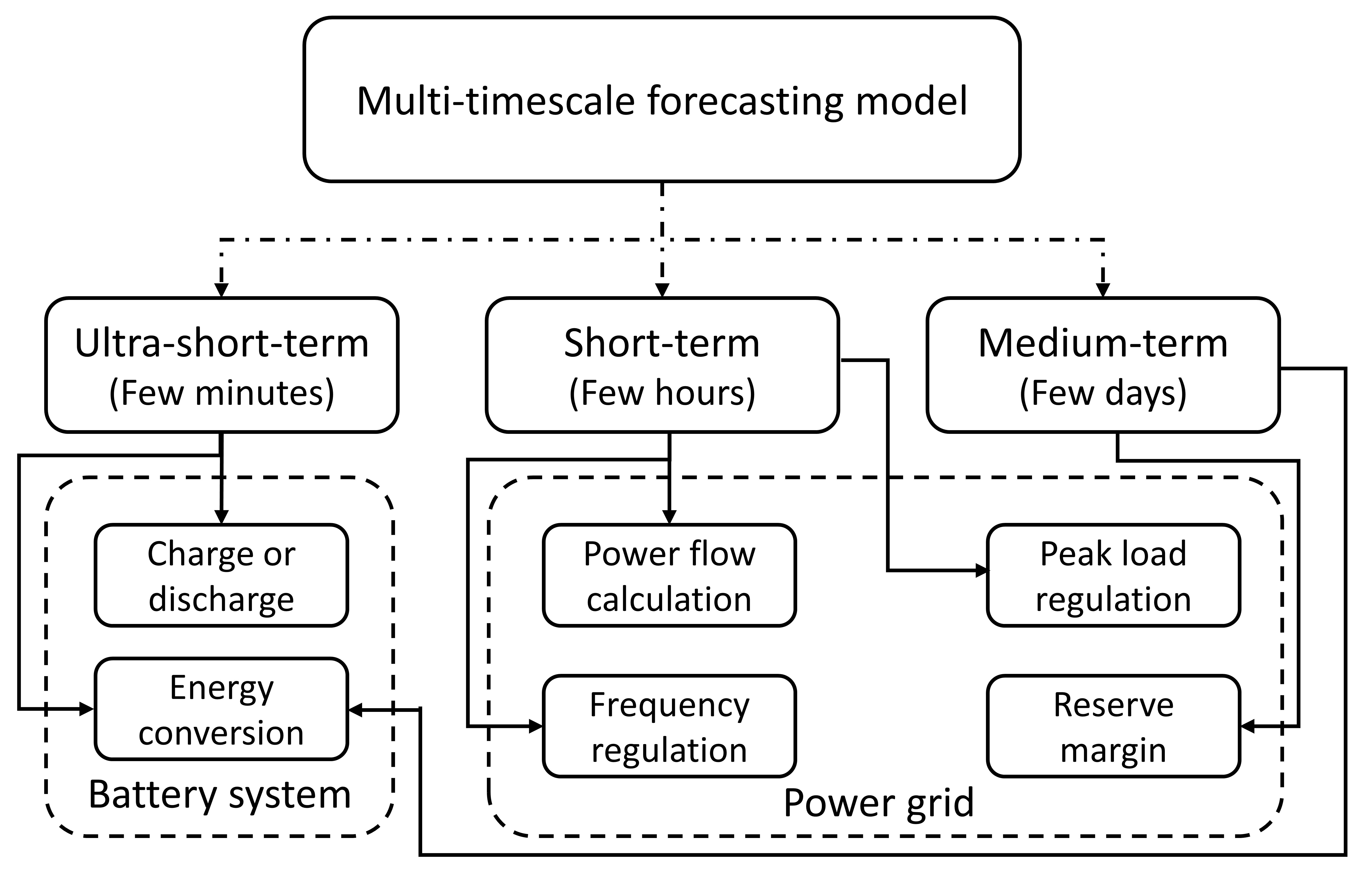}
  \caption{Schematic diagram of multi-time scale forecasting}
  \label{fig:mul}
\end{figure}

The power system comprises a battery system and a grid. Wind energy generation can be input into these components to facilitate intelligent grid dispatch and surplus energy storage. The proposed multi-time scale forecasting framework in this study enables wind power forecasting across the ultra-short-term, short-term, and medium-term intervals, providing decision support for battery systems and grids. Figure \ref{fig:mul} depicts the simplified flowchart of the framework.

The power generation within a few minutes in the ultra-short-term forecast is influenced by the battery system.The battery system regulates its charging or discharging based on the projected power generation capacity connected to the grid. Additionally, it converts surplus energy into alternative sources to address the fundamental instability of wind power prior to its integration with the grid. Short-term forecasting analyzes wind power generation for multiple hours, with a particular emphasis on the control problem of the power grid (Power flow calculation, frequency regulation and etc.) and the problem of intraday regulation. It not only ensures the stability of wind power grid connection, but also manages the misalignment between wind energy and load peaks.The medium-term forecast optimizes the balance between backup energy (typically traditional sources like coal) and wind energy in the grid through multi-day wind power generation forecasts.Additionally,it enables planning for the allocation of materials necessary for energy conversion in the energy system.

The schematic diagram of the proposed framework in this paper is shown in Figure \ref{fig:fra}.The framework comprises five primary components: data acquisition, data preprocessing, TPE algorithm, VMD decomposition, and model structure. The steps of the TPE-VMD-TFT method prediction process are as follows:

\begin{figure}[hbt!]
  \centering
  \includegraphics[width=1\textwidth]{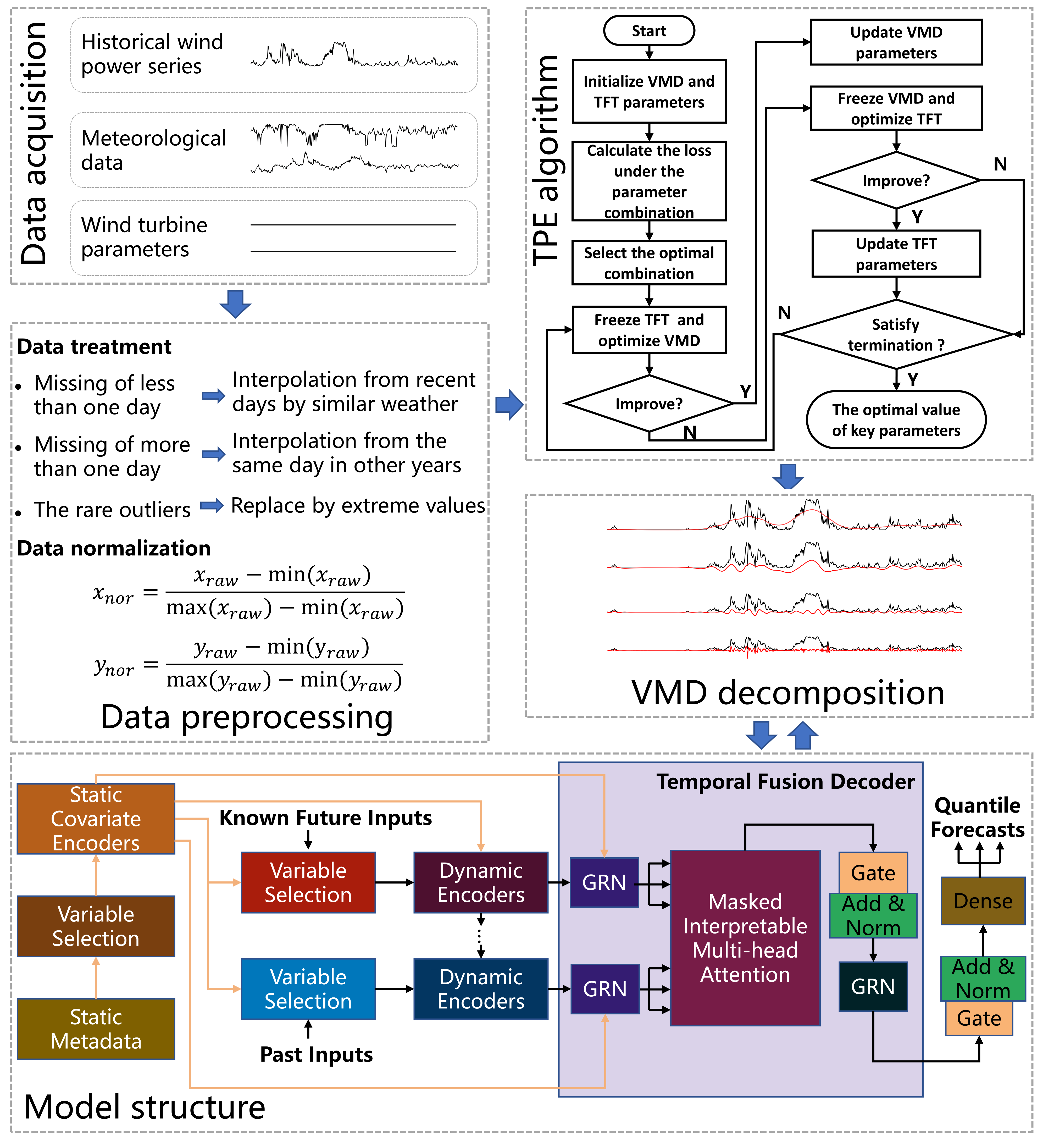}
  \caption{Structural of the proposed framework.}
  \label{fig:fra}
\end{figure}

\begin{itemize}
  \item [Step 1:]
  Real wind power generation data from wind power farms, including historical wind power series, meteorological data, and wind turbine parameters, are obtained.
  \item [Step 2:]
  The missing values are filled through interpolation, and outliers are rectified. The data is then normalized using the min-max method. 
  \item [Step 3:]
  The TPE algorithm can be used to optimize the key parameters of the VMD algorithm and the TFT model. First initialize the parameters of VMD and TFT, then calculate the loss and select the optimal parameter combination. The next step is tuning the parameters, freeze the parameters of the TFT and optimize the parameters of the VMD. If the prediction result improves, go to the next step and update the parameters of the VMD. If there is no improvement, freeze the parameters of the VMD and optimize the parameters of the TFT. If the prediction result improves, go to the next step and update the parameters of TFT. Loop until there is no significant improvement, or the maximum number of iteration is exceeded. Therefore, the optimal parameters of VMD and TFT can be obtained through the TPE algorithm.
  \item [Step 4:]
  The VMD algorithm, optimized with parameter tuning, is employed for decomposing the wind power generation series. This decomposition procedure yields multiple subseries exhibiting diverse frequencies. Consequently, the original series' volatility is efficiently reduced, enabling the capture of crucial features.
  \item [Step 5:]
  The parameter-optimized TFT model takes the decomposed subseries, meteorological data, and wind turbine parameters as inputs. These input variables undergo classification within the TFT model. The historical wind power subseries and meteorological data are classified as dynamic time-varying variables, while wind turbine parameters are classified as static covariates. Classification is also applied to weeks, months, and seasons, categorizing them as dynamic time-invariant variables. This classification process facilitates the extraction of distinct data features and enhances the generalization performance of the model.
  \end{itemize}

\subsection{TPE algorithm}\label{subsec4}

For wind turbine data $X$, we can represent it as a series $X = \left\{ x_1,x_2,\cdots,x_n,\hat{x} \right\}^{l}$ with $l$ time steps. Where $\hat{x}$ is the active power, and $x_n$ are the other features like wind speed. For a decomposition algorithm $D$, it turns the forecasting target $\left\{ \hat{x} \right\}^{l}$ into multi series $\left\{ \hat{x}_1,\hat{x}_2,\cdots,\hat{x}_m \right\}^{l}$. The task of wind power forecasting model $f$ is to fit $m$ sub sets like $X_m = \left\{ x_1,x_2,\cdots,x_n,\hat{x}_m \right\}^{l}$ to give corresponding $m$ predictions $\left\{ \hat{y}_1,\hat{y}_2,\cdots,\hat{y}_m \right\}^{l}$, where the predicted active power $Y = \sum \left\{ \hat{y}_1,\hat{y}_2,\cdots,\hat{y}_m \right\}^{l} = \left\{ y \right\}^{l}$.

As shown in Algorithm \ref{algorithm:test}, the hyperparameter configurations for the decomposition algorithm $D$ and power forecasting model $f$ are denoted as $U$ and $V$, respectively. Initially, we initialize the groups $U$ and $V$ with $N_{init}$ elements to obtain the initial observation set denoted by  $O = \left\{(U_i, V_i) \right\}, i=1,2,\ldots,N_{init}$. The minimization goal is achieved by evaluating the model's loss on the test set. We utilize the TPE algorithm to optimize each subset and store the optimized records in $O$. The optimization process is terminated when the maximum number of attempts is reached.

\begin{algorithm}
  \caption{Decomposition and model optimization}
  \label{algorithm:test}
  \begin{algorithmic}[1]
  \Require{Decomposition algorithm $D$; Forecasting model $f$; Wind turbine data $X$; Loss threshold $\theta$}   
  \Ensure{$u,v = U,V {|}_{(U,V, \mathbb{L}) \in O}^{min} L$}             
  \State $ O \gets \varnothing $   
  \State for $i = 1,2, \ldots, N_{init}$ do:      
  \State \quad $Randomly \quad pick (U_i, V_i)$      \Comment{Initialization}   
  \State \quad $\mathbb{L} = nMAE(f_{V_i}(D_{U_i}(X)))$   
  \State \quad $ O \gets O \cup {(U_i, V_i, \mathbb{L})}$   
  \State for $i = 1,2, \ldots, N_{max}$ do       
  \State \quad $u,v = U,V {|}_{(U,V, \mathbb{L}) \in O}^{min} L$  \Comment{Find the $U,V$ when $L$ is the minimum value}         
  \State \quad $U^{'} = TPE( u, v, f_v(D_{u^{*}}(X)))$   \Comment{* is the place holder for target function}     
  \State \quad $Improve = nMAE(f_{V_i}(D_{U^{'}}(X))) - nMAE(f_v(D_u(X)))$       
  \State \quad if $Improve \textgreater \theta$ do:            
  \State \qquad $O \gets O \cup \left\{ (U^{'}, v, \mathbb{L}) \right\}$       
  \State \quad $V^{'} = TPE(u, v, f_{v^{*}}(D_u(X)))$       
  \State \quad $Improve = nMAE(f_{V^{'}}(D_U(X))) - nMAE(f_v(D_u(X)))$     
  \State \quad if $Improve \textgreater \theta$ do:           
  \State \qquad $O \gets O \cup \left\{ (u, V^{'}, \mathbb{L}) \right\}$         
  
  \end{algorithmic}
\end{algorithm}

\section{Data description and evaluation metrics}\label{sec5}
\subsection{Data description}\label{subsec5}

The dataset used in this study is obtained from the La Haute Borne wind farm located in Meuse, France, at a longitude of 5.6013 E and a latitude of 48.4503 N. Meuse is situated on the western coast and experiences an oceanic climate characterized by an average annual temperature of 11.36$^{\circ}$C. The region exhibits a narrow annual temperature range with limited temperature extremes, featuring warm summers and cool winters. It possesses abundant wind energy resources and experiences high wind speeds attributed to the influence of the atlantic sea breeze.

The dataset used in this study is the Engie wind dataset provided by an electricity company in France \cite{engie}. The dataset is derived from the Supervisory Control and Data Acquisition (SCADA) system and contains daily electricity production records from 2012 to 2018 for four wind power turbines with a 2 MW rating. The training data span from 2012 to 2015, the validation set uses data from 2016, and the test data consists of records from 2017 and 2018. Each wind turbine is associated with three features: power, wind turbine, and meteorology data.The power features include active power, reactive power, and apparent power. The wind turbine data consist of twenty-two parameters such as converter torque, generator converter speed, and pitch angle, among others. The meteorology features include outdoor temperature, wind speed, and absolute wind direction. Each feature is measured in terms of average, minimum, maximum, and standard deviation values. The dataset contains missing cells ranging from 0.02\% to 0.05\% and missing records from 0.65\% to 0.88\%. The missing values, which occur for periods shorter than one day, are imputed using data from nearby days with similar weather conditions. For rare outliers, such as temperatures below -10 degrees (which never occur in the Grand Est region), they are replaced with the corresponding extreme values. The original dataset has a data reading frequency of 10 minutes, which is resampled to a resolution of one hour due to the high wind data instability. The increased data density creates additional challenges in prediction.

\begin{figure}[hbt!]
\centering
\includegraphics[width=1\textwidth]{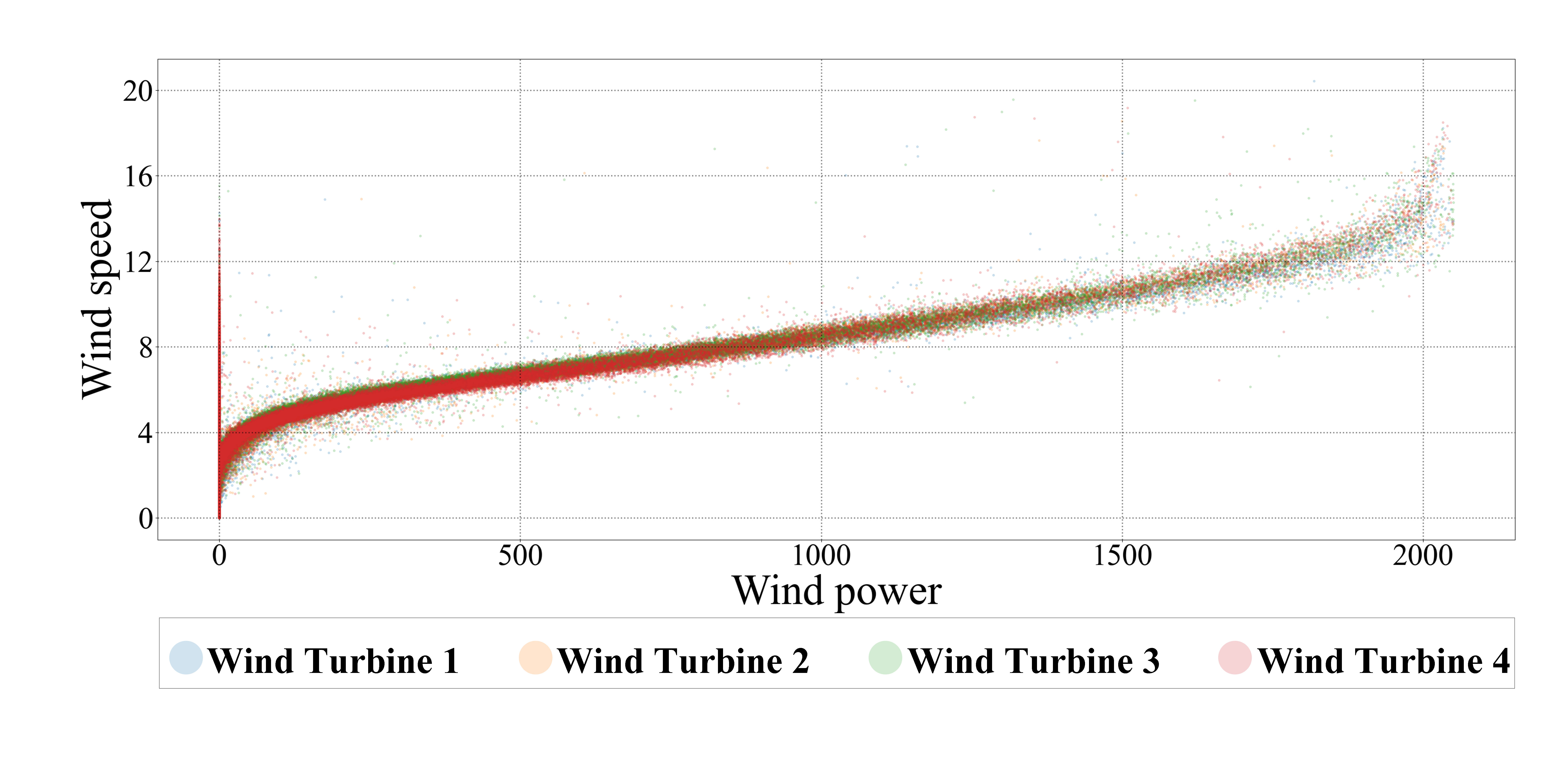}
\caption{Scatter plot of wind power and wind speed from raw data}
\label{fig:raw}
\end{figure}

\begin{figure}[hbt!]
\centering
\includegraphics[width=1\textwidth]{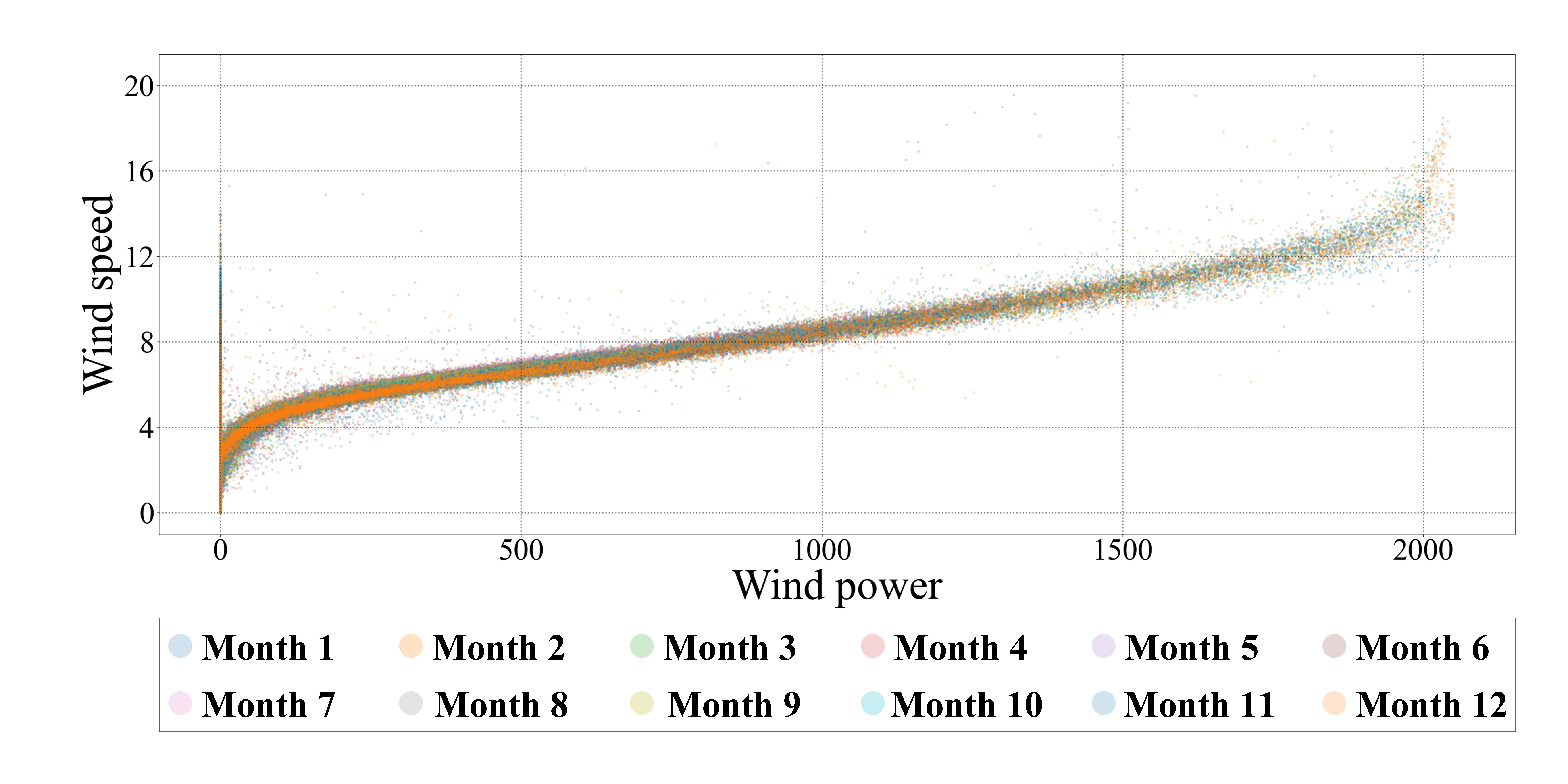}
\caption{Scatter plot of wind power and wind speed according to different months}
\label{fig:mon}
\end{figure}

The correlation between wind power production and meteorological variables varies depending on the location. Wind speed exhibits a stronger correlation with wind power production compared to other meteorological variables at the La Haute Borne wind farm. Figure \ref{fig:raw} displays scatter plots of wind power and wind speed using raw data collected from the four wind farms. Figure \ref{fig:mon} exhibits the scatter plot resulting from data segmentation by month. These figures demonstrate a gradual clustering of points along the diagonal, indicating a strong correlation between wind power and wind speed.

\subsection{Evaluation Metrics}\label{subsec5}

To assess the effectiveness of the proposed method, we employed two statistical metrics for prediction evaluation: nMAE and nRMSE. These metrics normalize the loss as percentages, which facilitates the comparison of error ranges among different wind turbines. Additionally, nMAE utilizes absolute error to prevent losses from being offset and exhibits low sensitivity to outliers. The definitions of nMAE and nRMSE are provided below.

\begin{equation}
  \label{eq:mae}
  nMAE = \frac{1}{Ty_{max}}\sum_{i=1}^T{|y_i - \hat{y}_i|}
\end{equation}

\begin{equation}
  \label{eq:rmse}
  nRMSE = \sqrt{\frac{1}{Ty_{max}^2} \sum_{i=1}^T(y_i - \hat{y}_i)^2}
\end{equation}

Here, $T$ represents the forecast length, $y_i$ signifies the actual wind power at point $i$, $\hat{y}_i$ denotes the predicted value of wind power at point $i$, and $y_{max}$ corresponds to the maximum value of the actual wind power. In studies on wind power forecasting, the units of RMSE and MAE can vary between kW, kWh, and percentage, depending on the forecasted variable. In this paper, each evaluation metric is presented in percentages to facilitate comparison with other wind power forecasting models.

\section{Experimental results and analysis}\label{subsec5}

The machine learning models use in this section were implemented using Python 3.8 and Torch 1.13.1, Pytorch-forecasting 0.10.2, and Pytorch-lightning 1.7.2. This paper uses historical data collected from the SCADA system of the Engie wind farm. All models utilize the "TimeSeriesDataSet" for data preprocessing and employ early stopping techniques to prevent overfitting. The performance of the proposed framework is validated using nMAE, nRMSE, and stability analysis.

\subsection{Multi-Time Scale Forecasting of automatic framework}

\begin{figure}[hbt!]
  \centering
  \includegraphics[width=1\textwidth]{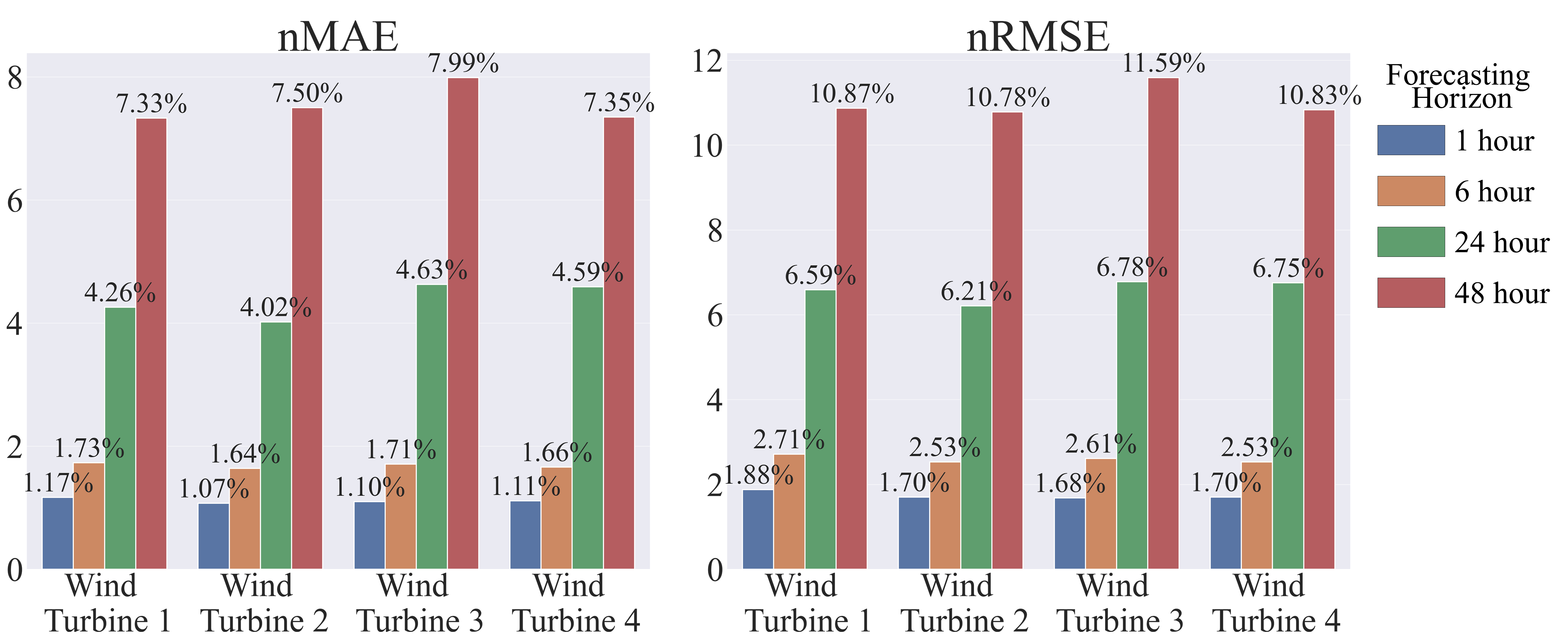}
  \caption{The results of multi-time scale prediction}
  \label{fig:fh}
  \end{figure}

The results of multi-time scale prediction for the four wind turbines, using the proposed automatic framework, are shown in Figure \ref{fig:fh}. It can be observed from the figure that the prediction error gaps for each wind turbine are small, and the overall prediction results are accurate, indicating the effectiveness of the proposed framework in predicting wind power generation at multi-time scale, which benefits the enhancement of the power system regulation.

\subsection{Forecasting performance comparison based on the time scale of 24h and 48h}

In this study, TPE-VMD-TFT exhibits better performance than other methods. Two prediction intervals of 24 hours and 48 hours are selected. Firstly, we compare TPE-VMD-TFT with individual models, and then we compare TPE-VMD-TFT with models based on the proposed framework. Furthermore, we verify the validity by analyzing the performance of TPE-VMD-TFT in different months, seasons, and years.

\subsubsection{Comparison of TPE-VMD-TFT and individual models }\label{subsubsec5}

ANN, LSTM, CNN-LSTM, RNN-LSTM, XGB, and TFT are commonly employed for wind power prediction. In this study, we compare them with the proposed TPE-VMD-TFT method, as shown in Table \ref{tab:res1}. In the TPE-VMD-TFT method, the parameters are optimized by VMD through TFT, which utilizes the data decomposed by VMD for training and prediction. The individual models utilize the original wind power time series data for training and prediction.

  \begin{sidewaystable}[]
    \caption{TPE-VMD-TFT method and other models for 24-h and 48-h ahead prediction error results.}
    \label{tab:res1}
    \begin{tabular*}{\textheight}{@{\extracolsep\fill}cccccccccc}
    \toprule
    \multirow{3}{*}{Model} & \multicolumn{4}{c}{Wind   Turbine 1} & \multirow{3}{*}{Model}          & \multicolumn{4}{c}{Wind   Turbine 2} \\ \cmidrule(lr){2-5} \cmidrule(l){7-10} 
                & \multicolumn{2}{c}{24-h} & \multicolumn{2}{c}{48-h} &             & \multicolumn{2}{c}{24-h} & \multicolumn{2}{c}{48-h} \\
                & nMAE        & nRMSE      & nMAE        & nRMSE      &             & nMAE        & nRMSE      & nMAE        & nRMSE      \\ \midrule
    ANN         & 14.34\%     & 18.31\%    & 18.42\%     & 21.64\%    & ANN         & 13.65\%     & 17.31\%    & 15.47\%     & 18.96\%    \\
    LSTM        & 12.94\%     & 17.95\%    & 17.02\%     & 27.00\%    & LSTM        & 11.89\%     & 16.79\%    & 16.11\%     & 25.14\%    \\
    CNN-LSTM    & 12.83\%     & 19.53\%    & 14.97\%     & 22.50\%    & CNN-LSTM    & 11.94\%     & 17.33\%    & 13.84\%     & 20.82\%    \\
    RNN-LSTM    & 12.73\%     & 18.02\%    & 15.15\%     & 22.27\%    & RNN-LSTM    & 12.20\%     & 16.94\%    & 14.01\%     & 20.56\%    \\
    XGB         & 12.63\%     & 17.58\%    & 15.10\%     & 19.97\%    & XGB         & 11.61\%     & 16.31\%    & 13.81\%     & 18.45\%    \\
    TFT         & 11.93\%     & 17.58\%    & 16.06\%     & 23.00\%    & TFT         & 11.41\%     & 16.74\%    & 14.28\%     & 20.04\%    \\
    TPE-VMD-TFT            & 4.26\%  & 6.59\%  & 7.33\% & 10.87\% & \multicolumn{1}{l}{TPE-VMD-TFT} & 4.02\%  & 6.21\%  & 7.50\% & 10.78\% \\ \midrule
    \multirow{3}{*}{Model} & \multicolumn{4}{c}{Wind Turbine 3}   & \multirow{3}{*}{Model}          & \multicolumn{4}{c}{Wind Turbine 4}   \\ \cmidrule(lr){2-5} \cmidrule(l){7-10} 
                & \multicolumn{2}{c}{24-h} & \multicolumn{2}{c}{48-h} &             & \multicolumn{2}{c}{24-h} & \multicolumn{2}{c}{48-h} \\
                & nMAE        & nRMSE      & nMAE        & nRMSE      &             & nMAE        & nRMSE      & nMAE        & nRMSE      \\ \midrule
    ANN         & 16.11\%     & 19.53\%    & 20.77\%     & 23.74\%    & ANN         & 17.42\%     & 20.30\%    & 19.55\%     & 22.53\%    \\
    LSTM        & 15.36\%     & 19.63\%    & 19.91\%     & 29.47\%    & LSTM        & 14.56\%     & 18.51\%    & 18.98\%     & 27.49\%    \\
    CNN-LSTM    & 14.94\%     & 19.65\%    & 16.73\%     & 23.62\%    & CNN-LSTM    & 14.05\%     & 18.90\%    & 17.80\%     & 28.13\%    \\
    RNN-LSTM    & 13.87\%     & 19.05\%    & 16.96\%     & 23.27\%    & RNN-LSTM    & 13.81\%     & 18.28\%    & 15.85\%     & 22.48\%    \\
    XGB         & 14.73\%     & 18.95\%    & 17.33\%     & 21.56\%    & XGB         & 13.47\%     & 18.08\%    & 15.60\%     & 20.19\%    \\
    TFT         & 15.78\%     & 20.20\%    & 17.15\%     & 23.51\%    & TFT         & 13.92\%     & 19.05\%    & 15.18\%     & 22.46\%    \\
    TPE-VMD-TFT & 4.63\%      & 6.78\%     & 7.99\%      & 11.59\%    & TPE-VMD-TFT & 4.59\%      & 6.75\%     & 7.35\%      & 10.83\%    \\ \bottomrule
    \end{tabular*}%
    \end{sidewaystable}
The table indicates that the TPE-VMD-TFT method has smaller nMAE and nRMSE values compared to the individual models for wind power data from all four turbines. In 24-hour ahead wind power forecasting, the TPE-VMD-TFT method has nMAE values of 4.26\%, 4.02\%, 4.63\%, and 4.59\%.  The prediction errors at wind turbine 1 for the ANN, LSTM, CNN-LSTM, RNN-LSTM, XGB, and TFT models are 70.29\%, 67.08\%, 66.80\%, 66.54\%, 66.27\%, and 64.29\% higher than the errors of the TPE-VMD-TFT method, respectively. Similarly, the nRMSE values for the TPE-VMD-TFT method are 6.59\%, 6.21\%, 6.78\%, and 6.75\%, and the prediction errors at wind turbine 1 for the individual models are 64.01\%, 63.29\%, 66.26\%, 63.43\%, 62.51\%, and 62.51\% higher than the errors of the TPE-VMD-TFT method, respectively. For 48-hour ahead forecasting, the TPE-VMD-TFT method achieves nMAE values of 7.33\%, 7.5\%, 7.99\%, and 7.35\%. The prediction errors at wind turbine 1 for the ANN, LSTM, CNN-LSTM, RNN-LSTM, XGB, and TFT models are 60.21\%, 56.93\%, 51.04\%, 51.62\%, 51.46\%, and 54.36\% higher than the errors of the TPE-VMD-TFT method, respectively. Similarly, the nRMSE values for the TPE-VMD-TFT method are 10.87\%, 10.78\%, 11.59\%, and 10.83\%, and the prediction errors at wind turbine 1 for the individual models are 49.77\%, 59.74\%, 51.69\%, 51.19\%, 45.57\%, and 52.74\% higher than the errors of the TPE-VMD-TFT method, respectively. The superior performance of the TPE-VMD-TFT method can be attributed to its use of an optimized VMD to decompose the wind power series, resulting in smoother IMFs compared to the original data, and reducing the impact of noise on the prediction results (subsection 5.3). Therefore, in terms of overall results, the TPE-VMD-TFT method significantly outperforms the individual models.

\begin{figure}[hbt!]
\centering
\includegraphics[width=1\textwidth]{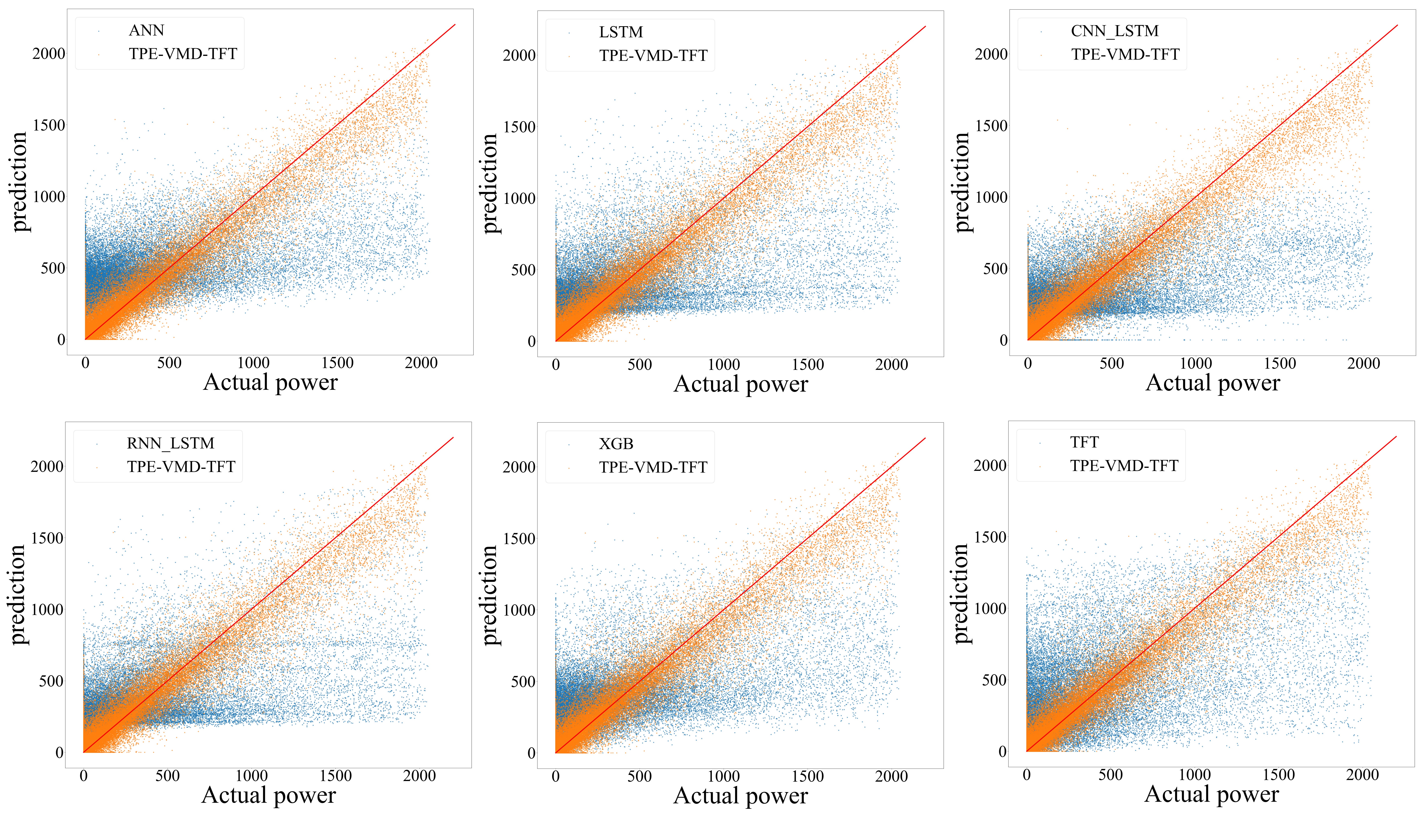}
\caption{Scatter plot of the models' predicted and actual values on four wind turbines.}
\label{fig:pre_act}
\end{figure}

We create scatter plots to compare the prediction performance of the TPE-VMD-TFT method and individual models on each day in the test set. For instance, Figure \ref{fig:pre_act} displays the predicted and actual values for the TPE-VMD-TFT method and other models on the four wind turbines in a 24-hour prediction. In the plot, the horizon axis represents the actual value of wind power, while the vertical axis represents the model's predicted value. The red diagonal line represents the ideal prediction, the yellow dots indicate the TPE-VMD-TFT method, and the blue dots represent the other models. The figure clearly shows that the yellow points are mainly concentrated around the diagonal line, whereas the blue points are more scattered. This indicates that the proposed TPE-VMD-TFT method has a higher accuracy rate compared to the individual models. Notably, the ANN and XGB models tend to predict points located above the diagonal line and closer to the origin of the coordinates, indicating a bias towards higher predictions for lower power. On the other hand, the LSTM, CNN-LSTM, and RNN-LSTM models exhibit a cluster of prediction points distributed in a region parallel to the horizon axis, suggesting that they predict a similar range of values regardless of the actual value. The scatter plot does not reveal any significant aggregation areas for the TFT model alone, and its points in high power display more dispersion. This demonstrates that the ANN, LSTM, CNN-LSTM, RNN-LSTM, and XGB models generate fixed output patterns to conform to the training metrics in the complex raw wind power data, whereas the TFT and XGB models do not demonstrate this pattern. Additionally, the individual models show very few points in the upper right region of the scatter plot, indicating poor prediction of peak power. Conversely, the TPE-VMD-TFT method displays prediction points distributed evenly on both sides of the ideal prediction, with a higher concentration of points in the peak region and less dispersion in the upper right point compared to the lower left. In comparison, the TPE-VMD-TFT method performs best in predicting low power compared to the other individual models. Furthermore, although the prediction error is higher at high power for the TPE-VMD-TFT method, it still represents a significant improvement over the other models.

\subsubsection{Comparison of TPE-VMD-TFT and models based on the automatic framework} \label{subsubsec5}

The proposed TPE-VMD-TFT model outperformed all other individual models discussed in the previous subsection. Additionally, the accuracy of the prediction is significantly impacted by the decomposition algorithm. For instance, the TPE-VMD-TFT model achieves higher prediction accuracy compared to the TFT model mentioned earlier. Therefore, the optimized decomposition algorithm used in our TPE-VMD-TFT model can be applied to other individual models for training and prediction purposes. In this subsection, we conduct a comparative analysis of TPE-VMD-TFT with other individual models using the proposed framework. The comparison includes all the decomposition algorithms and models mentioned before, namely: TPE-EEMD-ANN, TPE-EEMD-LSTM, TPE-EEMD-CNN-LSTM, TPE-EEMD-RNN-LSTM, TPE-EEMD-XGB, TPE-EEMD-TFT, TPE-VMD-ANN, TPE-VMD-LSTM, TPE-VMD-CNN-LSTM, TPE-VMD-RNN-LSTM, TPE-VMD-XGB, TPE-MSTL-ANN, TPE-MSTL-LSTM, TPE-MSTL-CNN-LSTM, TPE-MSTL-RNN-LSTM, TPE-MSTL-XGB, and TPE-MSTL-TFT.

  \begin{sidewaystable}[]
    \caption{Prediction results of TPE-VMD-TFT and models based on proposed frame.}
    \label{tab:res2}
    \begin{tabular*}{\textheight}{@{}cccccccccccccllllll@{}}
    \cmidrule(r){1-13}
     &
      \multicolumn{12}{c}{Wind   Turbine 1} &
       &
       &
       &
       &
       &
       \\
     &
       &
      \multicolumn{4}{c}{24-h} &
       &
       &
      \multicolumn{4}{c}{48-h} &
       &
       &
       &
       &
       &
       &
       \\ \cmidrule(lr){3-6} \cmidrule(lr){9-12}
     &
      \multicolumn{2}{c}{TPE-EEMD} &
      \multicolumn{2}{c}{TPE-VMD} &
      \multicolumn{2}{c}{TPE-MSTL} &
      \multicolumn{2}{c}{TPE-EEMD} &
      \multicolumn{2}{c}{TPE-VMD} &
      \multicolumn{2}{c}{TPE-MSTL} &
       &
       &
       &
       &
       &
       \\
    \multirow{-4}{*}{Model} &
      nMAE &
      nRMSE &
      nMAE &
      nRMSE &
      nMAE &
      nRMSE &
      nMAE &
      nRMSE &
      nMAE &
      nRMSE &
      nMAE &
      nRMSE &
       &
       &
       &
       &
       &
       \\ \cmidrule(r){1-13}
    ANN &
      7.57\% &
      10.65\% &
      7.27\% &
      9.37\% &
      \textbf{10.34\%} &
      \textbf{12.75\%} &
      16.55\% &
      20.35\% &
      15.97\% &
      18.51\% &
      17.28\% &
      20.06\% &
       &
       &
       &
       &
       &
       \\
    LSTM &
      7.75\% &
      10.84\% &
      5.98\% &
      8.96\% &
      11.66\% &
      15.84\% &
      16.34\% &
      25.56\% &
      17.61\% &
      27.37\% &
      17.61\% &
      27.31\% &
       &
       &
       &
       &
       &
       \\
    CNN-LSTM &
      8.60\% &
      12.29\% &
      6.23\% &
      9.22\% &
      11.13\% &
      16.11\% &
      14.85\% &
      22.18\% &
      14.66\% &
      22.12\% &
      14.87\% &
      22.70\% &
       &
       &
       &
       &
       &
       \\
    RNN-LSTM &
      8.75\% &
      12.76\% &
      7.07\% &
      9.78\% &
      12.05\% &
      15.48\% &
      15.62\% &
      21.95\% &
      15.20\% &
      22.22\% &
      15.11\% &
      22.30\% &
       &
       &
       &
       &
       &
       \\
    XGB &
      8.41\% &
      12.34\% &
      7.17\% &
      10.87\% &
      11.92\% &
      16.46\% &
      10.63\% &
      14.98\% &
      10.23\% &
      14.80\% &
      14.66\% &
      19.37\% &
       &
       &
       &
       &
       &
       \\
    TFT &
      \textbf{6.82\%} &
      \textbf{10.15\%} &
      {\color[HTML]{FF0000} \textbf{4.26\%}} &
      {\color[HTML]{FF0000} \textbf{6.59\%}} &
      10.68\% &
      15.31\% &
      \textbf{8.95\%} &
      \textbf{12.82\%} &
      {\color[HTML]{FF0000} \textbf{7.33\%}} &
      {\color[HTML]{FF0000} \textbf{10.87\%}} &
      \textbf{13.19\%} &
      \textbf{19.41\%} &
       &
       &
       &
       &
       &
       \\ \cmidrule(r){1-13}
     &
      \multicolumn{12}{c}{Wind Turbine 2} &
       &
       &
       &
       &
       &
       \\
     &
       &
      \multicolumn{4}{c}{24-h} &
       &
       &
      \multicolumn{4}{c}{48-h} &
       &
      \multicolumn{1}{c}{} &
      \multicolumn{1}{c}{} &
      \multicolumn{1}{c}{} &
      \multicolumn{1}{c}{} &
      \multicolumn{1}{c}{} &
      \multicolumn{1}{c}{} \\ \cmidrule(lr){3-6} \cmidrule(lr){9-12}
     &
      \multicolumn{2}{c}{TPE-EEMD} &
      \multicolumn{2}{c}{TPE-VMD} &
      \multicolumn{2}{c}{TPE-MSTL} &
      \multicolumn{2}{c}{TPE-EEMD} &
      \multicolumn{2}{c}{TPE-VMD} &
      \multicolumn{2}{c}{TPE-MSTL} &
       &
       &
       &
       &
       &
       \\
    \multirow{-4}{*}{Model} &
      nMAE &
      nRMSE &
      nMAE &
      nRMSE &
      nMAE &
      nRMSE &
      nMAE &
      nRMSE &
      nMAE &
      nRMSE &
      nMAE &
      nRMSE &
       &
       &
       &
       &
       &
       \\ \cmidrule(r){1-13}
    ANN &
      7.72\% &
      10.40\% &
      6.27\% &
      8.26\% &
      \textbf{9.39\%} &
      \textbf{11.45\%} &
      13.82\% &
      18.20\% &
      10.38\% &
      13.34\% &
      16.02\% &
      18.61\% &
       &
       &
       &
       &
       &
       \\
    LSTM &
      7.46\% &
      10.28\% &
      6.08\% &
      8.89\% &
      10.27\% &
      15.07\% &
      15.92\% &
      24.63\% &
      17.68\% &
      26.13\% &
      18.32\% &
      26.61\% &
       &
       &
       &
       &
       &
       \\
    CNN-LSTM &
      7.84\% &
      11.13\% &
      5.73\% &
      8.33\% &
      10.97\% &
      15.01\% &
      13.91\% &
      20.72\% &
      13.54\% &
      20.40\% &
      13.83\% &
      20.93\% &
       &
       &
       &
       &
       &
       \\
    RNN-LSTM &
      8.39\% &
      12.25\% &
      5.91\% &
      8.43\% &
      12.27\% &
      15.50\% &
      14.47\% &
      20.20\% &
      14.12\% &
      20.41\% &
      14.47\% &
      20.26\% &
       &
       &
       &
       &
       &
       \\
    XGB &
      7.62\% &
      11.29\% &
      6.83\% &
      10.21\% &
      11.10\% &
      15.49\% &
      \textbf{9.74\%} &
      \textbf{13.85\%} &
      9.64\% &
      13.71\% &
      14.01\% &
      18.49\% &
       &
       &
       &
       &
       &
       \\
    TFT &
      \textbf{6.55\%} &
      \textbf{9.55\%} &
      {\color[HTML]{FF0000} \textbf{4.02\%}} &
      {\color[HTML]{FF0000} \textbf{6.21\%}} &
      9.41\% &
      14.13\% &
      9.89\% &
      14.10\% &
      {\color[HTML]{FF0000} \textbf{7.50\%}} &
      {\color[HTML]{FF0000} \textbf{10.78\%}} &
      \textbf{12.12\%} &
      \textbf{17.35\%} &
       &
       &
       &
       &
       &
       \\ \cmidrule(r){1-13}
     &
      \multicolumn{12}{c}{Wind Turbine 3} &
       &
       &
       &
       &
       &
       \\
     &
       &
      \multicolumn{4}{c}{24-h} &
       &
       &
      \multicolumn{4}{c}{48-h} &
       &
      \multicolumn{1}{c}{} &
      \multicolumn{1}{c}{} &
      \multicolumn{1}{c}{} &
      \multicolumn{1}{c}{} &
      \multicolumn{1}{c}{} &
      \multicolumn{1}{c}{} \\ \cmidrule(lr){3-6} \cmidrule(lr){9-12}
     &
      \multicolumn{2}{c}{TPE-EEMD} &
      \multicolumn{2}{c}{TPE-VMD} &
      \multicolumn{2}{c}{TPE-MSTL} &
      \multicolumn{2}{c}{TPE-EEMD} &
      \multicolumn{2}{c}{TPE-VMD} &
      \multicolumn{2}{c}{TPE-MSTL} &
       &
       &
       &
       &
       &
       \\
    \multirow{-4}{*}{Model} &
      nMAE &
      nRMSE &
      nMAE &
      nRMSE &
      nMAE &
      nRMSE &
      nMAE &
      nRMSE &
      nMAE &
      nRMSE &
      nMAE &
      nRMSE &
       &
       &
       &
       &
       &
       \\ \cmidrule(r){1-13}
    ANN &
      10.49\% &
      13.16\% &
      10.87\% &
      12.91\% &
      11.12\% &
      \textbf{13.58\%} &
      22.43\% &
      25.33\% &
      17.71\% &
      20.23\% &
      19.44\% &
      22.47\% &
       &
       &
       &
       &
       &
       \\
    LSTM &
      9.29\% &
      11.76\% &
      7.74\% &
      10.21\% &
      11.99\% &
      16.79\% &
      19.75\% &
      29.02\% &
      19.66\% &
      29.05\% &
      20.03\% &
      29.52\% &
       &
       &
       &
       &
       &
       \\
    CNN-LSTM &
      9.23\% &
      12.60\% &
      8.00\% &
      10.29\% &
      12.90\% &
      17.07\% &
      16.04\% &
      22.59\% &
      16.50\% &
      23.55\% &
      16.63\% &
      23.81\% &
       &
       &
       &
       &
       &
       \\
    RNN-LSTM &
      9.36\% &
      13.22\% &
      9.08\% &
      11.04\% &
      13.26\% &
      16.64\% &
      17.82\% &
      22.92\% &
      17.24\% &
      23.12\% &
      17.07\% &
      23.18\% &
       &
       &
       &
       &
       &
       \\
    XGB &
      9.56\% &
      13.01\% &
      8.17\% &
      11.36\% &
      12.65\% &
      16.93\% &
      12.02\% &
      15.80\% &
      11.83\% &
      15.56\% &
      15.82\% &
      \textbf{20.18\%} &
       &
       &
       &
       &
       &
       \\
    TFT &
      \textbf{7.62\%} &
      \textbf{11.06\%} &
      {\color[HTML]{FF0000} \textbf{4.63\%}} &
      {\color[HTML]{FF0000} \textbf{6.78\%}} &
      \textbf{10.59\%} &
      15.13\% &
      \textbf{10.25\%} &
      \textbf{13.73\%} &
      {\color[HTML]{FF0000} \textbf{7.99\%}} &
      {\color[HTML]{FF0000} \textbf{11.59\%}} &
      \textbf{14.95\%} &
      21.04\% &
       &
       &
       &
       &
       &
       \\ \cmidrule(r){1-13}
     &
      \multicolumn{12}{c}{Wind Turbine 4} &
       &
       &
       &
       &
       &
       \\
     &
       &
      \multicolumn{4}{c}{24-h} &
       &
       &
      \multicolumn{4}{c}{48-h} &
       &
      \multicolumn{1}{c}{} &
      \multicolumn{1}{c}{} &
      \multicolumn{1}{c}{} &
      \multicolumn{1}{c}{} &
      \multicolumn{1}{c}{} &
      \multicolumn{1}{c}{} \\ \cmidrule(lr){3-6} \cmidrule(lr){9-12}
     &
      \multicolumn{2}{c}{TPE-EEMD} &
      \multicolumn{2}{c}{TPE-VMD} &
      \multicolumn{2}{c}{TPE-MSTL} &
      \multicolumn{2}{c}{TPE-EEMD} &
      \multicolumn{2}{c}{TPE-VMD} &
      \multicolumn{2}{c}{TPE-MSTL} &
       &
       &
       &
       &
       &
       \\
    \multirow{-4}{*}{Model} &
      nMAE &
      nRMSE &
      nMAE &
      nRMSE &
      nMAE &
      nRMSE &
      nMAE &
      nRMSE &
      nMAE &
      nRMSE &
      nMAE &
      nRMSE &
       &
       &
       &
       &
       &
       \\ \cmidrule(lr){2-13}
    ANN &
      10.37\% &
      12.79\% &
      7.98\% &
      10.07\% &
      10.62\% &
      \textbf{13.24\%} &
      18.20\% &
      21.51\% &
      13.36\% &
      16.16\% &
      17.59\% &
      20.30\% &
       &
       &
       &
       &
       &
       \\
    LSTM &
      8.00\% &
      11.10\% &
      6.48\% &
      9.57\% &
      11.09\% &
      15.72\% &
      17.72\% &
      27.11\% &
      17.43\% &
      27.08\% &
      17.55\% &
      27.26\% &
       &
       &
       &
       &
       &
       \\
    CNN-LSTM &
      8.79\% &
      12.39\% &
      6.88\% &
      9.66\% &
      10.91\% &
      15.33\% &
      15.62\% &
      23.19\% &
      15.51\% &
      22.63\% &
      15.66\% &
      22.70\% &
       &
       &
       &
       &
       &
       \\
    RNN-LSTM &
      8.98\% &
      13.00\% &
      7.21\% &
      9.72\% &
      11.68\% &
      15.25\% &
      16.66\% &
      22.12\% &
      16.13\% &
      22.26\% &
      15.93\% &
      22.40\% &
       &
       &
       &
       &
       &
       \\
    XGB &
      8.81\% &
      12.62\% &
      7.66\% &
      11.12\% &
      11.90\% &
      16.35\% &
      11.16\% &
      15.33\% &
      10.91\% &
      15.03\% &
      14.77\% &
      \textbf{19.33\%} &
       &
       &
       &
       &
       &
       \\
    TFT &
      \textbf{7.12\%} &
      \textbf{10.59\%} &
      {\color[HTML]{FF0000} \textbf{4.59\%}} &
      {\color[HTML]{FF0000} \textbf{6.75\%}} &
      \textbf{9.83\%} &
      14.64\% &
      \textbf{9.28\%} &
      \textbf{13.16\%} &
      {\color[HTML]{FF0000} \textbf{7.35\%}} &
      {\color[HTML]{FF0000} \textbf{10.83\%}} &
      \textbf{13.99\%} &
      20.34\% &
       &
       &
       &
       &
       &
       \\ \cmidrule(r){1-13}
    \end{tabular*}%
    \end{sidewaystable}

Table \ref{tab:res2} presents the results of the TPE-VMD-TFT model alongside other individual models, based on the proposed framework, for both 24-hour and 48-hour ahead wind power forecasts. The lowest values are obtained from the TPE-VMD-TFT model, and the TFT model performs the best among the various decomposition algorithms. This indicates that the TPE-VMD-TFT method outperforms both the TPE-EEMD-TFT and TPE-MSTL-TFT models for wind power data from all four turbines. Specifically, for wind turbine 1 predictions, the TPE-VMD-TFT method exhibits a 32.17\% and 60.11\% decrease in nMAE, as well as a 35.07\% and 56.96\% decrease in nRMSE, compared to the TPE-EEMD-TFT and TPE-MSTL-TFT models, respectively, for 24-hour ahead predictions. For 48-hour ahead predictions, the TPE-VMD-TFT method yields a decrease of 18.10\% and 44.72\% in the predicted wind power, and a decrease of 15.21\% and 42.43\% in nRMSE, compared to the TPE-EEMD-TFT and TPE-MSTL-TFT models, respectively. The aforementioned analysis elucidates the discrepancies in the predictive performance of the TFT model when combined with the three models generated by the proposed framework, with the TPE-VMD-TFT method exhibiting lower prediction errors.

The decomposition methods have varying impacts on the proposed framework across individual models. The VMD decomposition algorithm demonstrates superior performance in predicting 24-hour ahead wind power generation compared to EEMD and MSTL. Specifically, for the TPE-VMD-CNN-LSTM model on wind turbine 1 data, the nMAE value of 6.23\% represents a decrease of 27.56\% and 44.03\% compared to the TPE-EEMD-ANN and TPE-MSTL-ANN models, respectively. Similarly, the nRMSE value of 9.22\% indicates a decrease of 24.98\% and 42.77\%, respectively. Although most VMD decomposition algorithms outperform EEMD and MSTL in predicting 48-hour ahead wind power, some models exhibit discrepancies in their predictions. For instance, the TPE-VMD-LSTM model performs worse than the TPE-EEMD-LSTM model in terms of nMAE and nRMSE values for Wind Turbine 1 and 2 data. Consequently, determining the optimal decomposition method for a given dataset becomes challenging. Therefore, our proposed algorithm can automatically identify the most suitable parameters for other datasets and holds higher expectations for achieving high performance compared to methods that fix the decomposition algorithm and model.

\begin{figure}[hbt!]
\centering
\includegraphics[width=1\textwidth]{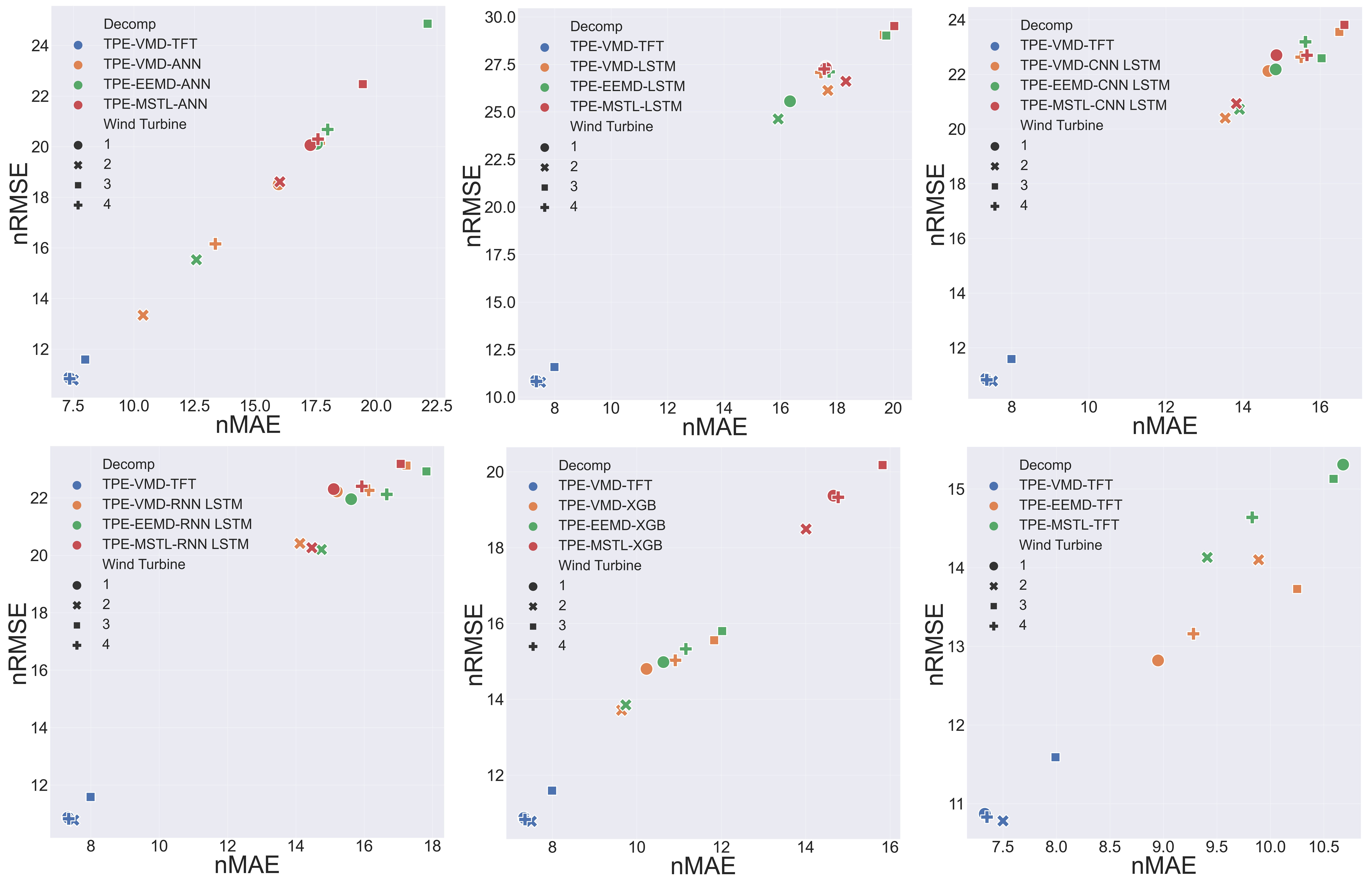}
\caption{The scatter plot of the model's 24-h ahead forecast results.}
\label{fig:sca_24}
\end{figure}

\begin{figure}[hbt!]
\centering
\includegraphics[width=1\textwidth]{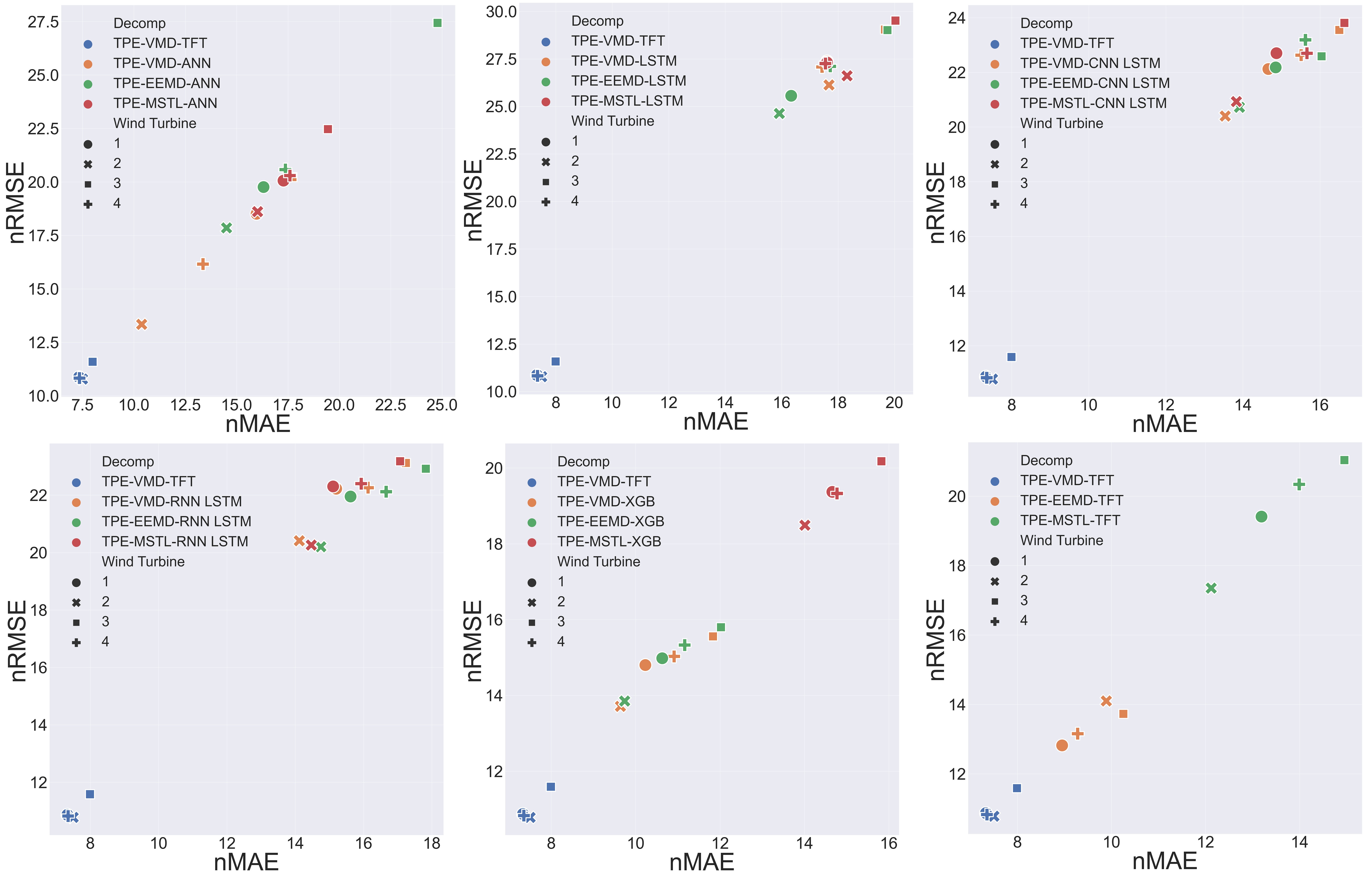}
\caption{The scatter plot of the model's 48-h ahead forecast results.}
\label{fig:sca_48}
\end{figure}

From the results presented in Table \ref{tab:res2} and the preceding analysis, it can be concluded that TPE-VMD-TFT is the most effective among the models studied. However, making a direct comparison between them is not straightforward due to the wide range of models and decomposition algorithms used. Consequently, we conduct separate comparisons between TPE-VMD-TFT and each class of models, including TPE-VMD-ANN, TPE-EEMD-ANN, and TPE-MSTL-ANN. Scatter plots illustrating the results for the 24-hour and 48-hour forecasts are displayed in Figures \ref{fig:sca_24} and \ref{fig:sca_48}, respectively. The x-axis represents the nMAE value of the model, while the y-axis represents the nRMSE value. To simplify the description, we refer to any model created by combining the proposed framework with a particular decomposition algorithm as TPE-Decomp-Model. For example, TPE-Decomp-TFT refers to TPE-VMD-TFT, TPE-EEMD-TFT, and TPE-MSTL-TFT.

As depicted in the figures, the blue points representing TPE-VMD-TFT are clustered near the origin of the scatter plot for all four wind turbine datasets. This clustering suggests that TPE-VMD-TFT outperforms the other composite models, indicating its superior predictive capabilities, particularly for volatile wind power data. The scatter plots for TPE-Decomp-LSTM, TPE-Decomp-RNN-LSTM, and TPE-Decomp-CNN-LSTM exhibit similar point distances across different decomposition algorithms, implying that these models are minimally affected by variations in decomposition methods. Moreover, the points for these models, as well as TPE-VMD-TFT, are distributed in the lower-left and upper-right corners of the scatter plot, implying a noticeable disparity between TPE-Decomp-LSTM, TPE-Decomp-RNN-LSTM, TPE-Decomp-CNN-LSTM, and TPE-VMD-TFT. On the other hand, the distributions of points for TPE-Decomp-ANN, TPE-Decomp-XGB, and TPE-Decomp-TFT are more dispersed along the diagonal. It can be observed that VMD performs better than EEMD, and EEMD performs better than MSTL. Therefore, the choice of decomposition algorithm significantly impacts the models, with VMD performing best across most of them.

\begin{sidewaystable}[]
\caption{Error reduction of the best model over the second best model}
\label{tab:error}
\begin{tabular*}{\textheight}{@{\extracolsep\fill}cccccccccc}
\toprule
\multirow{2}{*}{Target}                                         & Wind Turbine 1 & Wind Turbine 2 & Wind Turbine 3 & Wind Turbine 4 & Average \\ \cmidrule(l){2-6} 
                                                                & \multicolumn{5}{c}{24-h}                                                    \\ \midrule
\begin{tabular}[c]{@{}c@{}}nMAE\\ (\%Improvement)\end{tabular}  & 28.76\%        & 29.84\%        & 39.24\%        & 29.17\%        & 31.75\% \\
\begin{tabular}[c]{@{}c@{}}nRMSE\\ (\%Improvement)\end{tabular} & 26.45\%        & 25.45\%        & 33.59\%        & 29.47\%        & 28.74\% \\ \midrule
\multicolumn{1}{l}{}                                            & \multicolumn{5}{c}{48-h}                                                    \\ \cmidrule(l){2-6} 
\begin{tabular}[c]{@{}c@{}}nMAE\\ (\%Improvement)\end{tabular}  & 18.10\%        & 22.20\%        & 22.05\%        & 20.80\%        & 20.79\% \\
\begin{tabular}[c]{@{}c@{}}nRMSE\\ (\%Improvement)\end{tabular} & 15.21\%        & 19.19\%        & 15.59\%        & 17.71\%        & 16.93\% \\ \bottomrule
\end{tabular*}%
\end{sidewaystable}

We present the improvements in nMAE and nRMSE of TPE-VMD-TFT compared to the second-best model in Table \ref{tab:error}. The second-best model is determined separately for each wind turbine based on nMAE or nRMSE. In the 24-hour ahead forecasting, the TPE-VMD-TFT method exhibits reductions of 28.76\%, 29.84\%, 39.24\%, and 29.17\% compared to the second-best model for the four wind turbines, with an average reduction of 31.75\%. Regarding nRMSE, the TPE-VMD-TFT approach achieves an average reduction of 28.74\% compared to the second-best model. Similarly, in the 48-hour ahead forecasting, the TPE-VMD-TFT method achieves average reductions of 20.79\% and 16.93\% in nMAE and nRMSE, respectively, when compared to the second-best model.

\subsubsection{Classification analysis based on TPE-VMD-TFT}

In sub-subsection 5.4.1, we conduct a comparative analysis between the proposed TPE-VMD-TFT method and commonly used models for wind power forecasting. The results indicate that TPE-VMD-TFT demonstrates significant advantages over the other models. Subsequently, we apply TPE-DTBM to the remaining models, resulting in the creation of 17 additional models. The results consistently demonstrate that the proposed TPE-VMD-TFT method outperforms these models as well. To further analyze and confirm the validity of the TPE-VMD-TFT method, we group the forecasts based on month, season, and year.

\begin{figure}[hbt!]
  \centering
  \includegraphics[width=0.9\textwidth]{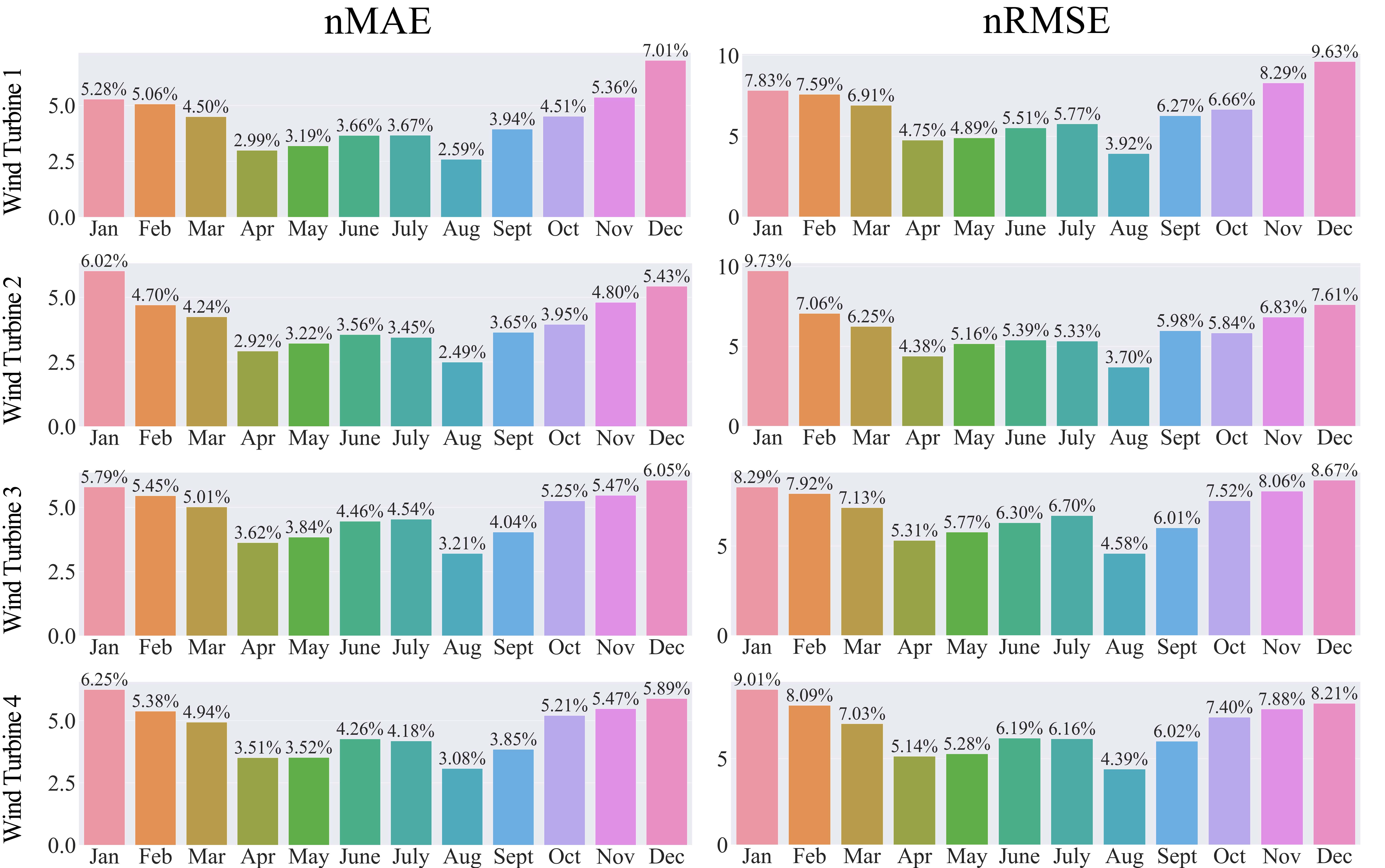}
  \caption{Monthly results of 24-h ahead forecasting based on TPE-VMD-TFT method (nMAE).}
  \label{fig:mon_24}
  \end{figure}

  \begin{figure}[hbt!]
  \centering
  \includegraphics[width=0.9\textwidth]{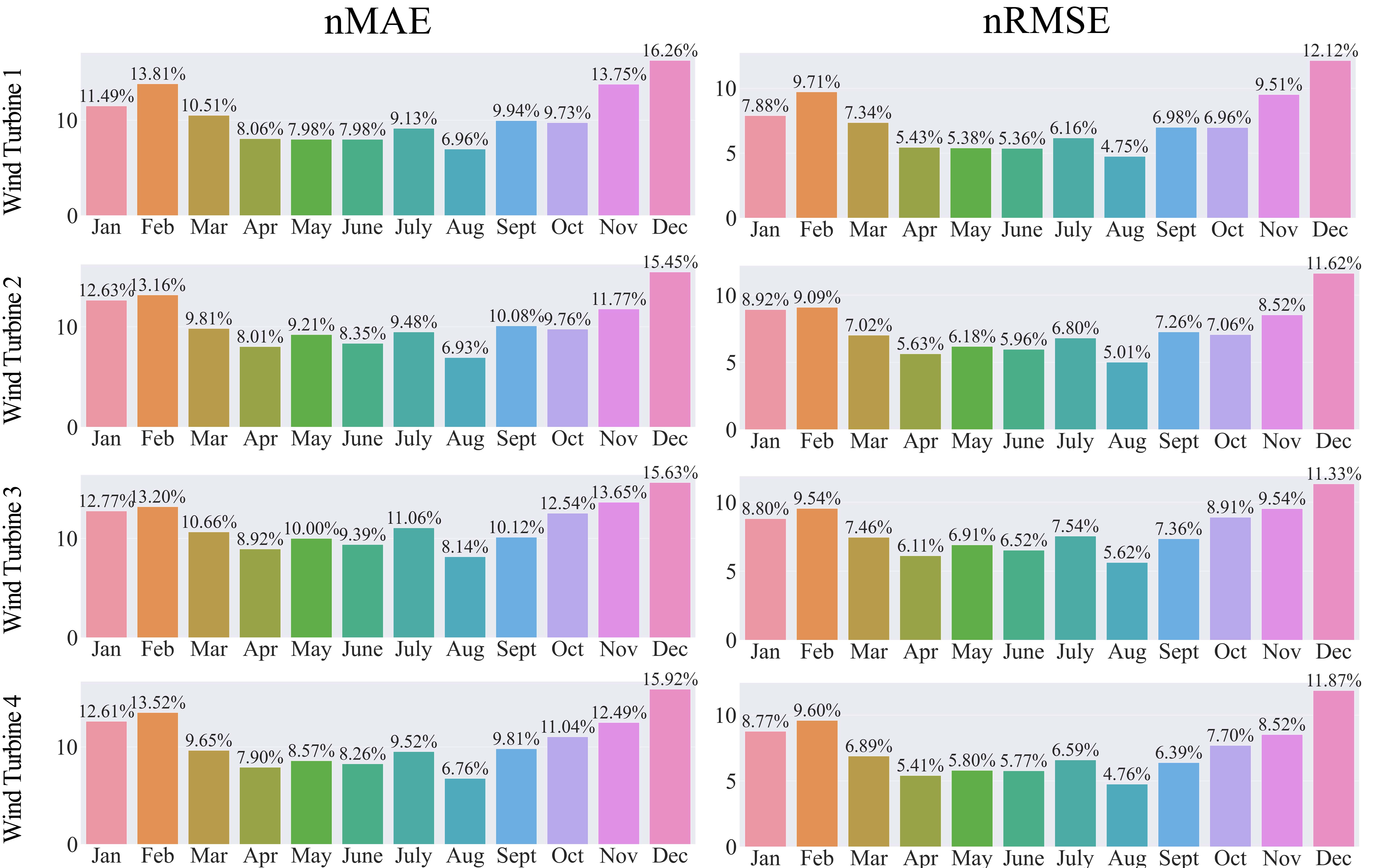}
  \caption{Monthly results of 48-h ahead forecasting based on TPE-VMD-TFT method (nMAE).}
  \label{fig:mon_48}
  \end{figure}

Figures \ref{fig:mon_24} and \ref{fig:mon_48} present the results of the TPE-VMD-TFT method for wind power forecasting, for 24-hour and 48-hour forecast horizons, organized by month. The horizontal axis displays the months from January to December, while the vertical axis indicates the forecast error. From a closer examination of figures \ref{fig:mon_24} and \ref{fig:mon_48}, it is evident that the nMAE values for the four wind turbines fluctuate between 2\% and 8\% throughout the year in the predicted 24-hour ahead wind power results. However, it is worth noting that wind turbine 1 exhibits an nMAE value of 7.01\% for December, while all other months display an nMAE value of less than 7\%. Furthermore, the forecast errors were significantly lower from April to September, but notably higher from January to March and October to December. Specifically, the error values for the four wind turbines were recorded as 2.59\%, 2.49\%, 3.21\%, and 3.08\% in August, which are comparatively smaller than the error values for the remaining months. Regarding the predicted 48-hour ahead wind power generation, the nMAE values for the four wind turbines ranged from 4\% to 13\% throughout the year. Notably, the forecast errors for December are relatively higher, with nMAE values surpassing 11\%. Conversely, the forecast errors for March to September exhibit relative consistency, with all nMAE values remaining below 8\%.

\begin{figure}[hbt!]
  \centering
  \includegraphics[width=\textwidth]{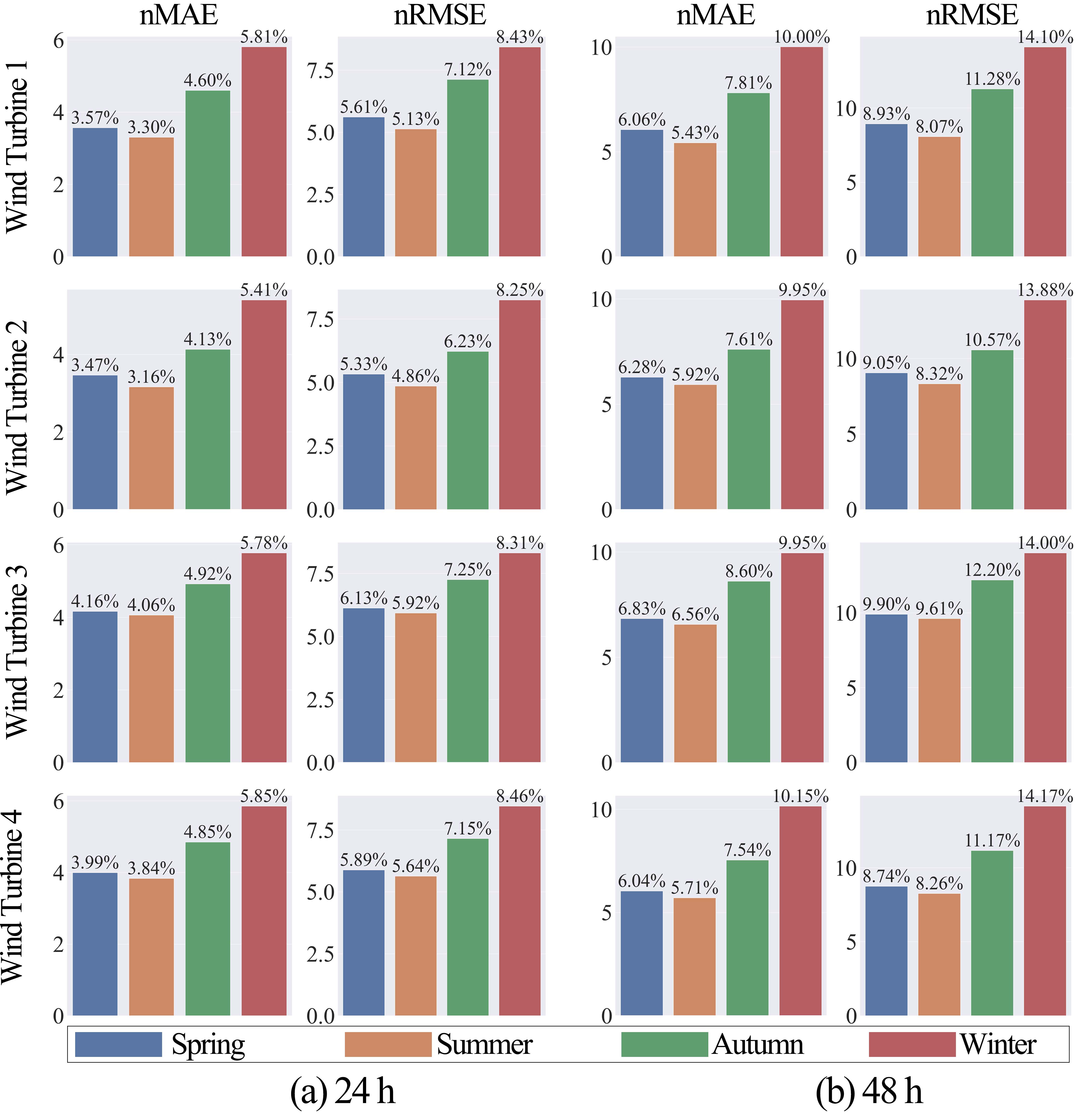}
  \caption{Seasonal results of 24-h and 48-h ahead forecasting based on TPE-VMD-TFT method (nMAE).}
  \label{fig:sea}
  \end{figure}

Figures \ref{fig:sea} display the results of the TPE-VMD-TFT method for wind power forecasting, specifically for 24-hour ahead and 48-hour ahead forecasts categorized by season. The horizontal axis represents the four seasons, while the vertical axis represents the prediction error. In terms of the wind farm's geographical location, the months of March, April, and May are classified as spring, June, July, and August as summer, September, October, and November as autumn, and December, January, and February as winter. As observed from the graph, the prediction error values for all four wind turbines remain consistently low across all seasons when forecasting 24 hours ahead, with nMAE values ranging from 3\% to 6\%. Interestingly, the summer months exhibit the lowest nMAE values, specifically measuring at 3.3\%, 3.16\%, 4.06\%, and 3.84\%. In contrast, the highest nMAE values occur during winter, recording values of 5.81\%, 5.41\%, 5.78\%, and 5.85\%. For the 48-hour ahead forecast, the nMAE values for the four wind turbines vary between 5\% and 11\% across all seasons, with relatively similar prediction errors in spring, summer, and autumn, and notably higher prediction errors in winter.

\begin{figure}[hbt!]
\centering
\includegraphics[width=\textwidth]{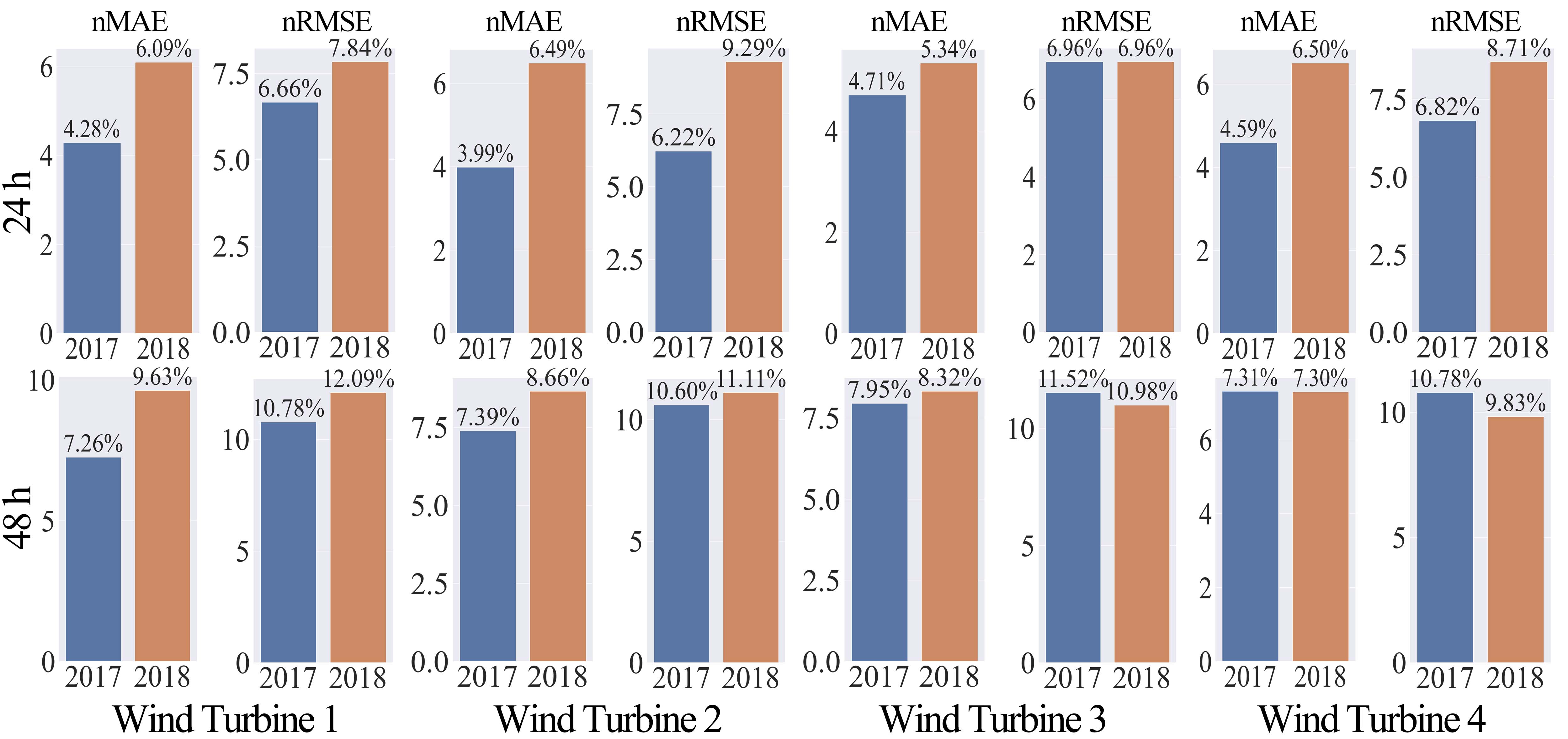}
\caption{Annual results of 24-h and 48-h ahead forecasting based on TPE-VMD-TFT method (nMAE).}
\label{fig:an}
\end{figure}

Figures \ref{fig:an} present the results of the TPE-VMD-TFT method for wind power forecasting, specifically for 24-hour ahead and 48-hour ahead forecasts categorized by year. The horizontal axis represents the different years, while the vertical axis represents the forecast error. From figures \ref{fig:an}, it is evident that when predicting wind power results for 24 hours ahead, all four wind turbines exhibit lower prediction errors in 2017 compared to 2018. Notably, wind turbine 2 demonstrates the most significant difference in predictions, with error values of 3.99\% and 6.49\% for 2017 and 2018, respectively. Conversely, when forecasting 48 hours ahead, the forecast errors for 2017 and 2018 are relatively similar. Wind turbine 4 records almost the same prediction error, while wind turbine 1 demonstrates a more substantial difference in prediction error, measuring at 7.26\% and 9.63\% for 2017 and 2018, respectively.

Based on the aforementioned analysis, it is evident that the TPE-VMD-TFT method demonstrates a commendable level of prediction accuracy across various months, seasons, and years. This finding implies the effectiveness of the proposed TPE-DTBM algorithm in enhancing the predictive performance of the TFT model.

\begin{figure}[hbt!]
\centering
\includegraphics[width=0.95\textwidth]{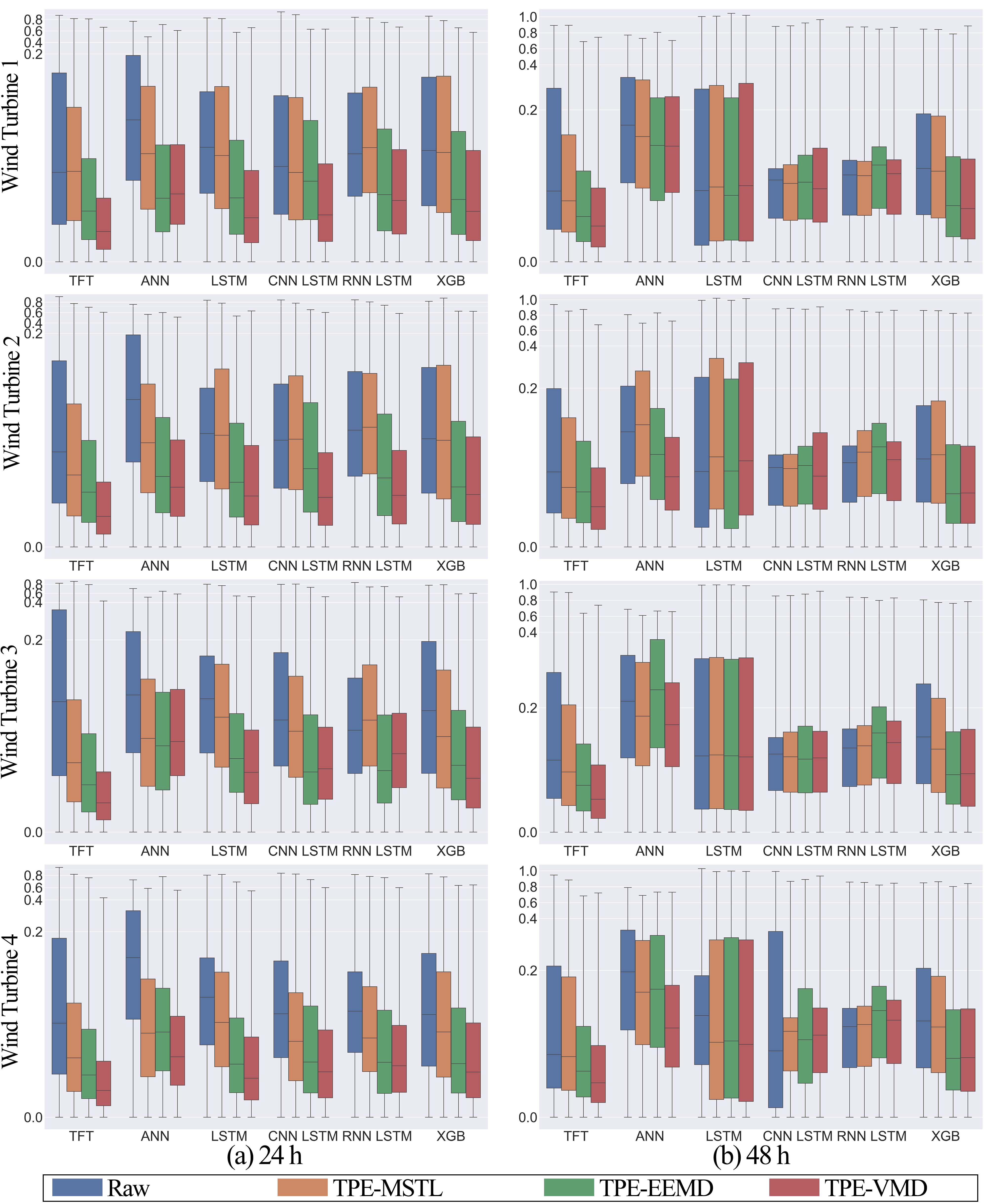}
\caption{Box plot of the error of the model for 24-h and 48-h ahead prediction.}
\label{fig:box}
\end{figure}

The figure \ref{fig:box} displays box plots that represent the prediction errors of all wind power forecasting models. The x-axis denotes the models used, while the y-axis indicates the distribution of prediction errors for these models. Given that the test set comprises 8832 time points, even with outliers ranging from 2\% to 10\%, there will still be a minimum of 170 outliers per method, which makes it challenging to visually represent these outliers with clarity. Thus, we choose to employ the Q1, Q2 and Q3 as the lower and upper boundaries of the box plot, respectively. The minimum and maximum values of the data will be used as the lower and upper boundaries of the whiskers, respectively. Upon observing the figure, the blue boxes correspond to the error distribution of individual models, the yellow boxes represent models based on the TPE-MSTL algorithm, the green boxes signify models based on the TPE-EEMD algorithm, and the red boxes indicate models based on the TPE-VMD algorithm. The graph reveals that the median values of the blue boxes are generally higher, and the boxes are longer. These findings suggest that the individual models exhibit less accuracy and stability in their predictions. Conversely, the red boxes typically exhibit lower median values and shorter boxes compared to the other colored boxes. These findings imply that the model based on the TPE-VMD algorithm demonstrates better predictive performance and greater stability than the models based on the TPE-EEMD and TPE-MSTL algorithms. Specifically, the TPE-VMD-TFT method exhibits the lowest median number of boxes and the shortest box length, thus indicating its superior performance compared to the other models.

\begin{figure}[hbt!]
\centering
\includegraphics[width=\textwidth]{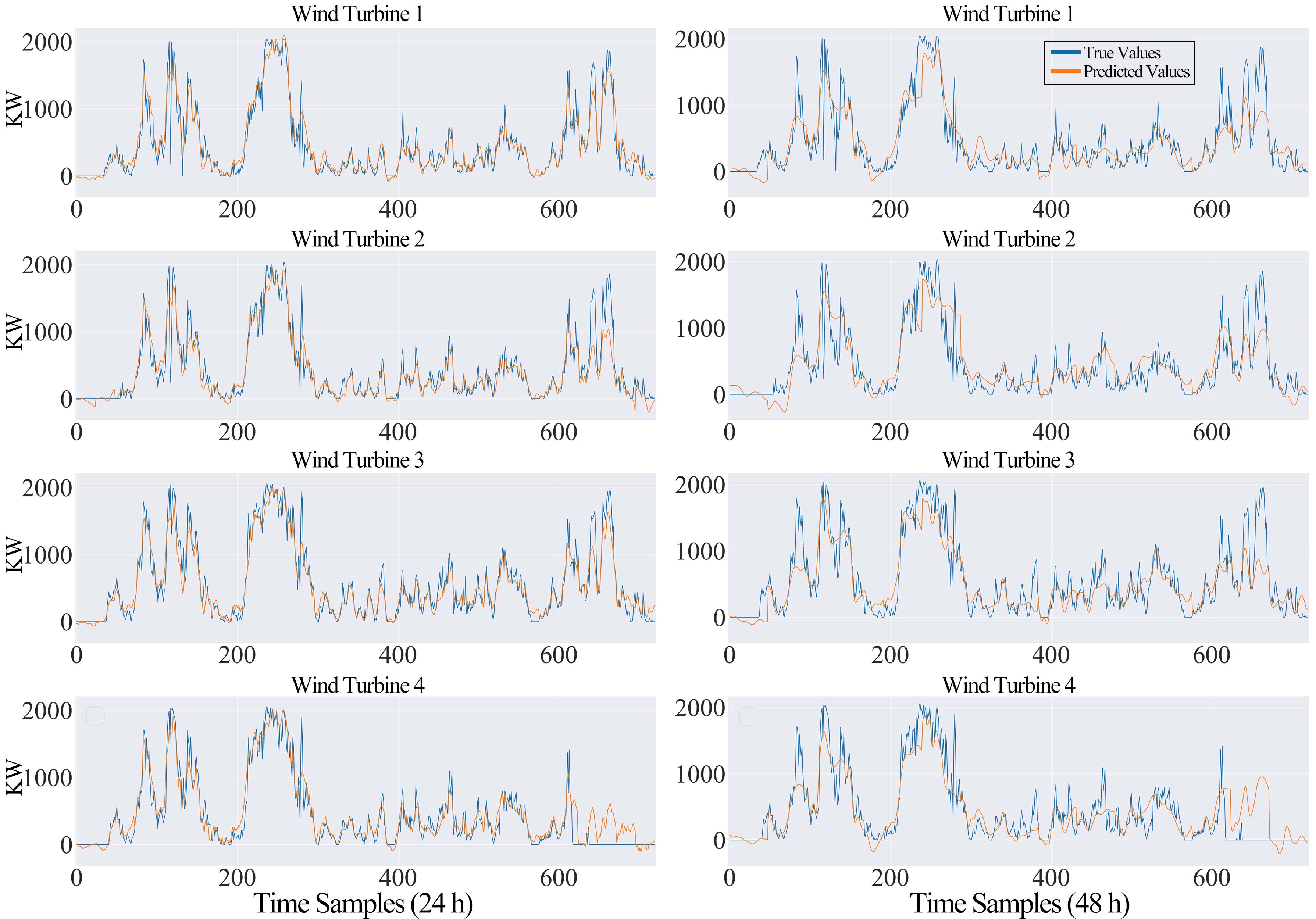}
\caption{24-h and 48-h ahead prediction results of the TPE-VMD-TFT method.}
\label{fig:pre_pro}
\end{figure}

To showcase the performance of the TPE-VMD-TFT method proposed in this study on wind power data, Figures \ref{fig:pre_pro} present the results of the TPE-VMD-TFT method for 24-hour and 48-hour ahead wind power predictions, respectively. In these figures, the actual and predicted values of wind power are denoted by the black and red lines, respectively. The graph reveals a greater overlap between the two lines when predicting 24 hours ahead, indicating that the predicted values of wind power align closely with the actual values. Despite a reduced prediction performance when forecasting 48 hours ahead, the results remain reasonable. This can be attributed to the increased complexity, instability, and uncertainty associated with predicting wind power further into the future. Overall, the TPE-VMD-TFT method demonstrates improved accuracy compared to individual models, as evidenced by the closer proximity of its prediction curves to the original wind power series for volatile wind power data. By decomposing the wind power series using the TPE-VMD framework to mitigate the impact of noise on prediction outcomes, relevant features are effectively extracted for the TFT model, resulting in improved prediction performance.

\section{Conclusion}

Accurate wind power forecasting is crucial for power system scheduling and planning. However, forecasting wind power is challenging due to its high uncertainty, discontinuity, and volatile fluctuations. In this study, we propose a novel medium-term forecasting framework, the TPE-VMD-TFT method, based on TPE and decomposition algorithms, for predicting wind power from wind turbines. Existing literature often relies on empirically selecting hyperparameters for the decomposition algorithm, resulting in suboptimal or inferior predictions. To address these issues, we introduce the TPE-DTBM algorithm, which serves as a vital component of the TPE-VMD-TFT approach and offers generalizability to other common decomposition algorithms and models used in wind power forecasting. To evaluate the performance of our proposed method, we conduct experiments using the benchmark Engie wind dataset from an electricity company in France, employing nMAE and nRMSE as evaluation metrics.

The nMAE values for wind power prediction 24 hours ahead using our proposed method were 4.2\%, 4.02\%, 4.63\%, and 4.59\% for the four wind turbines, respectively. Correspondingly, the nRMSE values were 6.59\%, 6.21\%, 6.78\%, and 6.75\%. For the 48-hour ahead predictions, the nMAE values were 7.33\%, 7.5\%, 7.99\%, and 7.35\%. Furthermore, the nRMSE values were 10.87\%, 10.78\%, 11.59\%, and 10.83\%. Our proposed method achieved a reduction of over 50\% in nMAE values and over 40\% in nRMSE values compared to other individual time series prediction models (e.g., ANN, LSTM, CNN-LSTM, RNN-LSTM, and XGB). Moreover, when compared to wind power generation forecasting methods that utilize different decomposition algorithms for data feature extraction, our proposed method exhibited superior forecasting performance. By utilizing the TPE-DTBM algorithm to optimize the VMD parameters for wind power series decomposition, the TPE-VMD-TFT method effectively extracts wind power characteristics and reduces historical data noise, leading to a significant improvement in prediction accuracy.

To further analyze and validate the efficacy of the TPE-VMD-TFT method, the forecasts were categorized based on month, season, and year. The experiments confirm that the proposed method exhibits high prediction accuracy across different months, seasons, and years, indicating the effectiveness of the TPE-DTBM algorithm in enhancing the prediction performance of the TFT model. Additionally, a box plot of the prediction errors of the wind power forecasting models illustrates the greater stability of the proposed method in comparison to other comparative models.

The wind power forecasting problem is addressed in this study through the introduction of a novel medium-term forecasting framework and the TPE-VMD-TFT method, which exhibit superior forecasting performance and generalization capability. Nevertheless, there remain some challenging issues that warrant further investigation. For instance, variations in meteorological data across different geographical locations have not been considered, potentially affecting the prediction accuracy. Moreover, the proposed framework for wind power prediction can be extended to other areas of energy prediction, such as photovoltaic power and wind speed forecasting, which will be the focus of our future endeavors.

\backmatter

\bmhead{Acknowledgments}

This research was supported by CAS "Light of West China" Program.









\bibliography{sn-bibliography}


\begin{thebibliography}{38}
\ifx \bisbn   \undefined \def \bisbn  #1{ISBN #1}\fi
\ifx \binits  \undefined \def \binits#1{#1}\fi
\ifx \bauthor  \undefined \def \bauthor#1{#1}\fi
\ifx \batitle  \undefined \def \batitle#1{#1}\fi
\ifx \bjtitle  \undefined \def \bjtitle#1{#1}\fi
\ifx \bvolume  \undefined \def \bvolume#1{\textbf{#1}}\fi
\ifx \byear  \undefined \def \byear#1{#1}\fi
\ifx \bissue  \undefined \def \bissue#1{#1}\fi
\ifx \bfpage  \undefined \def \bfpage#1{#1}\fi
\ifx \blpage  \undefined \def \blpage #1{#1}\fi
\ifx \burl  \undefined \def \burl#1{\textsf{#1}}\fi
\ifx \doiurl  \undefined \def \doiurl#1{\url{https://doi.org/#1}}\fi
\ifx \betal  \undefined \def \betal{\textit{et al.}}\fi
\ifx \binstitute  \undefined \def \binstitute#1{#1}\fi
\ifx \binstitutionaled  \undefined \def \binstitutionaled#1{#1}\fi
\ifx \bctitle  \undefined \def \bctitle#1{#1}\fi
\ifx \beditor  \undefined \def \beditor#1{#1}\fi
\ifx \bpublisher  \undefined \def \bpublisher#1{#1}\fi
\ifx \bbtitle  \undefined \def \bbtitle#1{#1}\fi
\ifx \bedition  \undefined \def \bedition#1{#1}\fi
\ifx \bseriesno  \undefined \def \bseriesno#1{#1}\fi
\ifx \blocation  \undefined \def \blocation#1{#1}\fi
\ifx \bsertitle  \undefined \def \bsertitle#1{#1}\fi
\ifx \bsnm \undefined \def \bsnm#1{#1}\fi
\ifx \bsuffix \undefined \def \bsuffix#1{#1}\fi
\ifx \bparticle \undefined \def \bparticle#1{#1}\fi
\ifx \barticle \undefined \def \barticle#1{#1}\fi
\bibcommenthead
\ifx \bconfdate \undefined \def \bconfdate #1{#1}\fi
\ifx \botherref \undefined \def \botherref #1{#1}\fi
\ifx \url \undefined \def \url#1{\textsf{#1}}\fi
\ifx \bchapter \undefined \def \bchapter#1{#1}\fi
\ifx \bbook \undefined \def \bbook#1{#1}\fi
\ifx \bcomment \undefined \def \bcomment#1{#1}\fi
\ifx \oauthor \undefined \def \oauthor#1{#1}\fi
\ifx \citeauthoryear \undefined \def \citeauthoryear#1{#1}\fi
\ifx \endbibitem  \undefined \def \endbibitem {}\fi
\ifx \bconflocation  \undefined \def \bconflocation#1{#1}\fi
\ifx \arxivurl  \undefined \def \arxivurl#1{\textsf{#1}}\fi
\csname PreBibitemsHook\endcsname

\bibitem[\protect\citeauthoryear{}{2023}]{un_2022}
\begin{botherref}
UN Climate Action
(2023).
\url{https://www.un.org/en/climatechange/science/causes-effects-climate-change}
\end{botherref}
\endbibitem

\bibitem[\protect\citeauthoryear{Wang et~al.}{2019}]{wang2019approaches}
\begin{barticle}
\bauthor{\bsnm{Wang}, \binits{Y.}},
\bauthor{\bsnm{Hu}, \binits{Q.}},
\bauthor{\bsnm{Li}, \binits{L.}},
\bauthor{\bsnm{Foley}, \binits{A.M.}},
\bauthor{\bsnm{Srinivasan}, \binits{D.}}:
\batitle{Approaches to wind power curve modeling: A review and discussion}.
\bjtitle{Renewable and Sustainable Energy Reviews}
\bvolume{116},
\bfpage{109422}
(\byear{2019})
\end{barticle}
\endbibitem

\bibitem[\protect\citeauthoryear{Luo et~al.}{2021}]{luo2021local}
\begin{barticle}
\bauthor{\bsnm{Luo}, \binits{L.}},
\bauthor{\bsnm{Zhuang}, \binits{Y.}},
\bauthor{\bsnm{Duan}, \binits{Q.}},
\bauthor{\bsnm{Dong}, \binits{L.}},
\bauthor{\bsnm{Yu}, \binits{Y.}},
\bauthor{\bsnm{Liu}, \binits{Y.}},
\bauthor{\bsnm{Chen}, \binits{K.}},
\bauthor{\bsnm{Gao}, \binits{X.}}:
\batitle{Local climatic and environmental effects of an onshore wind farm in
  north china}.
\bjtitle{Agricultural and Forest Meteorology}
\bvolume{308},
\bfpage{108607}
(\byear{2021})
\end{barticle}
\endbibitem

\bibitem[\protect\citeauthoryear{}{2022}]{gwec_2022}
\begin{botherref}
GWEC 2022
(2022).
\url{https://gwec.net/global-wind-report-2022/}
\end{botherref}
\endbibitem

\bibitem[\protect\citeauthoryear{Chang et~al.}{2014}]{chang2014literature}
\begin{barticle}
\bauthor{\bsnm{Chang}, \binits{W.-Y.}}, \betal:
\batitle{A literature review of wind forecasting methods}.
\bjtitle{Journal of Power and Energy Engineering}
\bvolume{2}(\bissue{04}),
\bfpage{161}
(\byear{2014})
\end{barticle}
\endbibitem

\bibitem[\protect\citeauthoryear{Aburiyana and
  El-Hawary}{2017}]{aburiyana2017overview}
\begin{bchapter}
\bauthor{\bsnm{Aburiyana}, \binits{G.}},
\bauthor{\bsnm{El-Hawary}, \binits{M.E.}}:
\bctitle{An overview of forecasting techniques for load, wind and solar
  powers}.
In: \bbtitle{2017 IEEE Electrical Power and Energy Conference (EPEC)},
pp. \bfpage{1}--\blpage{7}
(\byear{2017}).
\bcomment{IEEE}
\end{bchapter}
\endbibitem

\bibitem[\protect\citeauthoryear{Devi et~al.}{2020}]{devi2020hourly}
\begin{barticle}
\bauthor{\bsnm{Devi}, \binits{A.S.}},
\bauthor{\bsnm{Maragatham}, \binits{G.}},
\bauthor{\bsnm{Boopathi}, \binits{K.}},
\bauthor{\bsnm{Rangaraj}, \binits{A.}}:
\batitle{Hourly day-ahead wind power forecasting with the eemd-cso-lstm-efg
  deep learning technique}.
\bjtitle{Soft Computing}
\bvolume{24}(\bissue{16}),
\bfpage{12391}--\blpage{12411}
(\byear{2020})
\end{barticle}
\endbibitem

\bibitem[\protect\citeauthoryear{Hanifi et~al.}{2020}]{hanifi2020critical}
\begin{barticle}
\bauthor{\bsnm{Hanifi}, \binits{S.}},
\bauthor{\bsnm{Liu}, \binits{X.}},
\bauthor{\bsnm{Lin}, \binits{Z.}},
\bauthor{\bsnm{Lotfian}, \binits{S.}}:
\batitle{A critical review of wind power forecasting methods—past, present
  and future}.
\bjtitle{Energies}
\bvolume{13}(\bissue{15}),
\bfpage{3764}
(\byear{2020})
\end{barticle}
\endbibitem

\bibitem[\protect\citeauthoryear{Soman et~al.}{2010}]{soman2010review}
\begin{bchapter}
\bauthor{\bsnm{Soman}, \binits{S.S.}},
\bauthor{\bsnm{Zareipour}, \binits{H.}},
\bauthor{\bsnm{Malik}, \binits{O.}},
\bauthor{\bsnm{Mandal}, \binits{P.}}:
\bctitle{A review of wind power and wind speed forecasting methods with
  different time horizons}.
In: \bbtitle{North American Power Symposium 2010},
pp. \bfpage{1}--\blpage{8}
(\byear{2010}).
\bcomment{IEEE}
\end{bchapter}
\endbibitem

\bibitem[\protect\citeauthoryear{Wang et~al.}{2022}]{wang2022deep}
\begin{barticle}
\bauthor{\bsnm{Wang}, \binits{Y.}},
\bauthor{\bsnm{Xu}, \binits{H.}},
\bauthor{\bsnm{Zou}, \binits{R.}},
\bauthor{\bsnm{Zhang}, \binits{L.}},
\bauthor{\bsnm{Zhang}, \binits{F.}}:
\batitle{A deep asymmetric laplace neural network for deterministic and
  probabilistic wind power forecasting}.
\bjtitle{Renewable Energy}
\bvolume{196},
\bfpage{497}--\blpage{517}
(\byear{2022})
\end{barticle}
\endbibitem

\bibitem[\protect\citeauthoryear{Tian et~al.}{2022}]{tian2022developing}
\begin{barticle}
\bauthor{\bsnm{Tian}, \binits{C.}},
\bauthor{\bsnm{Niu}, \binits{T.}},
\bauthor{\bsnm{Wei}, \binits{W.}}:
\batitle{Developing a wind power forecasting system based on deep learning with
  attention mechanism}.
\bjtitle{Energy}
\bvolume{257},
\bfpage{124750}
(\byear{2022})
\end{barticle}
\endbibitem

\bibitem[\protect\citeauthoryear{Duan et~al.}{2022}]{duan2022novel}
\begin{barticle}
\bauthor{\bsnm{Duan}, \binits{J.}},
\bauthor{\bsnm{Wang}, \binits{P.}},
\bauthor{\bsnm{Ma}, \binits{W.}},
\bauthor{\bsnm{Fang}, \binits{S.}},
\bauthor{\bsnm{Hou}, \binits{Z.}}:
\batitle{A novel hybrid model based on nonlinear weighted combination for
  short-term wind power forecasting}.
\bjtitle{International Journal of Electrical Power \& Energy Systems}
\bvolume{134},
\bfpage{107452}
(\byear{2022})
\end{barticle}
\endbibitem

\bibitem[\protect\citeauthoryear{Xiong et~al.}{2022}]{xiong2022short}
\begin{barticle}
\bauthor{\bsnm{Xiong}, \binits{B.}},
\bauthor{\bsnm{Lou}, \binits{L.}},
\bauthor{\bsnm{Meng}, \binits{X.}},
\bauthor{\bsnm{Wang}, \binits{X.}},
\bauthor{\bsnm{Ma}, \binits{H.}},
\bauthor{\bsnm{Wang}, \binits{Z.}}:
\batitle{Short-term wind power forecasting based on attention mechanism and
  deep learning}.
\bjtitle{Electric Power Systems Research}
\bvolume{206},
\bfpage{107776}
(\byear{2022})
\end{barticle}
\endbibitem

\bibitem[\protect\citeauthoryear{Pei et~al.}{2022}]{pei2022short}
\begin{barticle}
\bauthor{\bsnm{Pei}, \binits{M.}},
\bauthor{\bsnm{Ye}, \binits{L.}},
\bauthor{\bsnm{Li}, \binits{Y.}},
\bauthor{\bsnm{Luo}, \binits{Y.}},
\bauthor{\bsnm{Song}, \binits{X.}},
\bauthor{\bsnm{Yu}, \binits{Y.}},
\bauthor{\bsnm{Zhao}, \binits{Y.}}:
\batitle{Short-term regional wind power forecasting based on spatial--temporal
  correlation and dynamic clustering model}.
\bjtitle{Energy Reports}
\bvolume{8},
\bfpage{10786}--\blpage{10802}
(\byear{2022})
\end{barticle}
\endbibitem

\bibitem[\protect\citeauthoryear{Li et~al.}{2022}]{li2022wind}
\begin{barticle}
\bauthor{\bsnm{Li}, \binits{J.}},
\bauthor{\bsnm{Zhang}, \binits{S.}},
\bauthor{\bsnm{Yang}, \binits{Z.}}:
\batitle{A wind power forecasting method based on optimized decomposition
  prediction and error correction}.
\bjtitle{Electric Power Systems Research}
\bvolume{208},
\bfpage{107886}
(\byear{2022})
\end{barticle}
\endbibitem

\bibitem[\protect\citeauthoryear{Rayi et~al.}{2022}]{rayi2022adaptive}
\begin{barticle}
\bauthor{\bsnm{Rayi}, \binits{V.K.}},
\bauthor{\bsnm{Mishra}, \binits{S.}},
\bauthor{\bsnm{Naik}, \binits{J.}},
\bauthor{\bsnm{Dash}, \binits{P.}}:
\batitle{Adaptive vmd based optimized deep learning mixed kernel elm
  autoencoder for single and multistep wind power forecasting}.
\bjtitle{Energy}
\bvolume{244},
\bfpage{122585}
(\byear{2022})
\end{barticle}
\endbibitem

\bibitem[\protect\citeauthoryear{Zhao et~al.}{2023}]{zhao2023hybrid}
\begin{barticle}
\bauthor{\bsnm{Zhao}, \binits{Z.}},
\bauthor{\bsnm{Yun}, \binits{S.}},
\bauthor{\bsnm{Jia}, \binits{L.}},
\bauthor{\bsnm{Guo}, \binits{J.}},
\bauthor{\bsnm{Meng}, \binits{Y.}},
\bauthor{\bsnm{He}, \binits{N.}},
\bauthor{\bsnm{Li}, \binits{X.}},
\bauthor{\bsnm{Shi}, \binits{J.}},
\bauthor{\bsnm{Yang}, \binits{L.}}:
\batitle{Hybrid vmd-cnn-gru-based model for short-term forecasting of wind
  power considering spatio-temporal features}.
\bjtitle{Engineering Applications of Artificial Intelligence}
\bvolume{121},
\bfpage{105982}
(\byear{2023})
\end{barticle}
\endbibitem

\bibitem[\protect\citeauthoryear{Gao et~al.}{2023}]{gao2023short}
\begin{barticle}
\bauthor{\bsnm{Gao}, \binits{X.}},
\bauthor{\bsnm{Guo}, \binits{W.}},
\bauthor{\bsnm{Mei}, \binits{C.}},
\bauthor{\bsnm{Sha}, \binits{J.}},
\bauthor{\bsnm{Guo}, \binits{Y.}},
\bauthor{\bsnm{Sun}, \binits{H.}}:
\batitle{Short-term wind power forecasting based on ssa-vmd-lstm}.
\bjtitle{Energy Reports}
\bvolume{9},
\bfpage{335}--\blpage{344}
(\byear{2023})
\end{barticle}
\endbibitem

\bibitem[\protect\citeauthoryear{Qiao et~al.}{2022}]{qiao2022wind}
\begin{barticle}
\bauthor{\bsnm{Qiao}, \binits{B.}},
\bauthor{\bsnm{Liu}, \binits{J.}},
\bauthor{\bsnm{Wu}, \binits{P.}},
\bauthor{\bsnm{Teng}, \binits{Y.}}:
\batitle{Wind power forecasting based on variational mode decomposition and
  high-order fuzzy cognitive maps}.
\bjtitle{Applied Soft Computing}
\bvolume{129},
\bfpage{109586}
(\byear{2022})
\end{barticle}
\endbibitem

\bibitem[\protect\citeauthoryear{Wang et~al.}{2021}]{wang2021review}
\begin{barticle}
\bauthor{\bsnm{Wang}, \binits{Y.}},
\bauthor{\bsnm{Zou}, \binits{R.}},
\bauthor{\bsnm{Liu}, \binits{F.}},
\bauthor{\bsnm{Zhang}, \binits{L.}},
\bauthor{\bsnm{Liu}, \binits{Q.}}:
\batitle{A review of wind speed and wind power forecasting with deep neural
  networks}.
\bjtitle{Applied Energy}
\bvolume{304},
\bfpage{117766}
(\byear{2021})
\end{barticle}
\endbibitem

\bibitem[\protect\citeauthoryear{Lazi{\'c} et~al.}{2010}]{lazic2010wind}
\begin{barticle}
\bauthor{\bsnm{Lazi{\'c}}, \binits{L.}},
\bauthor{\bsnm{Pejanovi{\'c}}, \binits{G.}},
\bauthor{\bsnm{{\v{Z}}ivkovi{\'c}}, \binits{M.}}:
\batitle{Wind forecasts for wind power generation using the eta model}.
\bjtitle{Renewable Energy}
\bvolume{35}(\bissue{6}),
\bfpage{1236}--\blpage{1243}
(\byear{2010})
\end{barticle}
\endbibitem

\bibitem[\protect\citeauthoryear{Yatiyana et~al.}{2017}]{yatiyana2017wind}
\begin{bchapter}
\bauthor{\bsnm{Yatiyana}, \binits{E.}},
\bauthor{\bsnm{Rajakaruna}, \binits{S.}},
\bauthor{\bsnm{Ghosh}, \binits{A.}}:
\bctitle{Wind speed and direction forecasting for wind power generation using
  arima model}.
In: \bbtitle{2017 Australasian Universities Power Engineering Conference
  (AUPEC)},
pp. \bfpage{1}--\blpage{6}
(\byear{2017}).
\bcomment{IEEE}
\end{bchapter}
\endbibitem

\bibitem[\protect\citeauthoryear{Firat et~al.}{2010}]{firat2010wind}
\begin{bchapter}
\bauthor{\bsnm{Firat}, \binits{U.}},
\bauthor{\bsnm{Engin}, \binits{S.N.}},
\bauthor{\bsnm{Saraclar}, \binits{M.}},
\bauthor{\bsnm{Ertuzun}, \binits{A.B.}}:
\bctitle{Wind speed forecasting based on second order blind identification and
  autoregressive model}.
In: \bbtitle{2010 Ninth International Conference on Machine Learning and
  Applications},
pp. \bfpage{686}--\blpage{691}
(\byear{2010}).
\bcomment{IEEE}
\end{bchapter}
\endbibitem

\bibitem[\protect\citeauthoryear{Bilal et~al.}{2018}]{bilal2018wind}
\begin{bchapter}
\bauthor{\bsnm{Bilal}, \binits{B.}},
\bauthor{\bsnm{Ndongo}, \binits{M.}},
\bauthor{\bsnm{Adjallah}, \binits{K.H.}},
\bauthor{\bsnm{Sava}, \binits{A.}},
\bauthor{\bsnm{Kebe}, \binits{C.M.}},
\bauthor{\bsnm{Ndiaye}, \binits{P.A.}},
\bauthor{\bsnm{Sambou}, \binits{V.}}:
\bctitle{Wind turbine power output prediction model design based on artificial
  neural networks and climatic spatiotemporal data}.
In: \bbtitle{2018 IEEE International Conference on Industrial Technology
  (ICIT)},
pp. \bfpage{1085}--\blpage{1092}
(\byear{2018}).
\bcomment{IEEE}
\end{bchapter}
\endbibitem

\bibitem[\protect\citeauthoryear{Jyothi and Rao}{2016}]{jyothi2016very}
\begin{bchapter}
\bauthor{\bsnm{Jyothi}, \binits{M.N.}},
\bauthor{\bsnm{Rao}, \binits{P.R.}}:
\bctitle{Very-short term wind power forecasting through adaptive wavelet neural
  network}.
In: \bbtitle{2016 Biennial International Conference on Power and Energy
  Systems: Towards Sustainable Energy (PESTSE)},
pp. \bfpage{1}--\blpage{6}
(\byear{2016}).
\bcomment{IEEE}
\end{bchapter}
\endbibitem

\bibitem[\protect\citeauthoryear{Wang et~al.}{2020}]{wang2020clustered}
\begin{barticle}
\bauthor{\bsnm{Wang}, \binits{Y.}},
\bauthor{\bsnm{Wang}, \binits{D.}},
\bauthor{\bsnm{Tang}, \binits{Y.}}:
\batitle{Clustered hybrid wind power prediction model based on arma, pso-svm,
  and clustering methods}.
\bjtitle{IEEE Access}
\bvolume{8},
\bfpage{17071}--\blpage{17079}
(\byear{2020})
\end{barticle}
\endbibitem

\bibitem[\protect\citeauthoryear{Shetty et~al.}{2016}]{shetty2016optimized}
\begin{bchapter}
\bauthor{\bsnm{Shetty}, \binits{R.P.}},
\bauthor{\bsnm{Sathyabhama}, \binits{A.}},
\bauthor{\bsnm{Rai}, \binits{A.A.}}, \betal:
\bctitle{Optimized radial basis function neural network model for wind power
  prediction}.
In: \bbtitle{2016 Second International Conference on Cognitive Computing and
  Information Processing (CCIP)},
pp. \bfpage{1}--\blpage{6}
(\byear{2016}).
\bcomment{IEEE}
\end{bchapter}
\endbibitem

\bibitem[\protect\citeauthoryear{Yan and Ouyang}{2019}]{yan2019advanced}
\begin{barticle}
\bauthor{\bsnm{Yan}, \binits{J.}},
\bauthor{\bsnm{Ouyang}, \binits{T.}}:
\batitle{Advanced wind power prediction based on data-driven error correction}.
\bjtitle{Energy conversion and management}
\bvolume{180},
\bfpage{302}--\blpage{311}
(\byear{2019})
\end{barticle}
\endbibitem

\bibitem[\protect\citeauthoryear{Ackermann and
  S{\"o}der}{2000}]{ackermann2000wind}
\begin{barticle}
\bauthor{\bsnm{Ackermann}, \binits{T.}},
\bauthor{\bsnm{S{\"o}der}, \binits{L.}}:
\batitle{Wind energy technology and current status: a review}.
\bjtitle{Renewable and sustainable energy reviews}
\bvolume{4}(\bissue{4}),
\bfpage{315}--\blpage{374}
(\byear{2000})
\end{barticle}
\endbibitem

\bibitem[\protect\citeauthoryear{Bashir}{2022}]{bashir2022principle}
\begin{barticle}
\bauthor{\bsnm{Bashir}, \binits{M.B.A.}}:
\batitle{Principle parameters and environmental impacts that affect the
  performance of wind turbine: an overview}.
\bjtitle{Arabian Journal for Science and Engineering}
\bvolume{47}(\bissue{7}),
\bfpage{7891}--\blpage{7909}
(\byear{2022})
\end{barticle}
\endbibitem

\bibitem[\protect\citeauthoryear{Hochreiter and
  Schmidhuber}{1997}]{hochreiter1997long}
\begin{barticle}
\bauthor{\bsnm{Hochreiter}, \binits{S.}},
\bauthor{\bsnm{Schmidhuber}, \binits{J.}}:
\batitle{Long short-term memory}.
\bjtitle{Neural computation}
\bvolume{9}(\bissue{8}),
\bfpage{1735}--\blpage{1780}
(\byear{1997})
\end{barticle}
\endbibitem

\bibitem[\protect\citeauthoryear{Chen and Guestrin}{2016}]{chen2016xgboost}
\begin{bchapter}
\bauthor{\bsnm{Chen}, \binits{T.}},
\bauthor{\bsnm{Guestrin}, \binits{C.}}:
\bctitle{Xgboost: A scalable tree boosting system}.
In: \bbtitle{Proceedings of the 22nd Acm Sigkdd International Conference on
  Knowledge Discovery and Data Mining},
pp. \bfpage{785}--\blpage{794}
(\byear{2016})
\end{bchapter}
\endbibitem

\bibitem[\protect\citeauthoryear{Lim et~al.}{2021}]{lim2021temporal}
\begin{barticle}
\bauthor{\bsnm{Lim}, \binits{B.}},
\bauthor{\bsnm{Ar{\i}k}, \binits{S.{\"O}.}},
\bauthor{\bsnm{Loeff}, \binits{N.}},
\bauthor{\bsnm{Pfister}, \binits{T.}}:
\batitle{Temporal fusion transformers for interpretable multi-horizon time
  series forecasting}.
\bjtitle{International Journal of Forecasting}
\bvolume{37}(\bissue{4}),
\bfpage{1748}--\blpage{1764}
(\byear{2021})
\end{barticle}
\endbibitem

\bibitem[\protect\citeauthoryear{Clevert et~al.}{2015}]{clevert2015fast}
\begin{botherref}
\oauthor{\bsnm{Clevert}, \binits{D.-A.}},
\oauthor{\bsnm{Unterthiner}, \binits{T.}},
\oauthor{\bsnm{Hochreiter}, \binits{S.}}:
Fast and accurate deep network learning by exponential linear units (elus).
arXiv preprint arXiv:1511.07289
(2015)
\end{botherref}
\endbibitem

\bibitem[\protect\citeauthoryear{Wu and Huang}{2009}]{wu2009ensemble}
\begin{barticle}
\bauthor{\bsnm{Wu}, \binits{Z.}},
\bauthor{\bsnm{Huang}, \binits{N.E.}}:
\batitle{Ensemble empirical mode decomposition: a noise-assisted data analysis
  method}.
\bjtitle{Advances in adaptive data analysis}
\bvolume{1}(\bissue{01}),
\bfpage{1}--\blpage{41}
(\byear{2009})
\end{barticle}
\endbibitem

\bibitem[\protect\citeauthoryear{Dragomiretskiy and
  Zosso}{2013}]{dragomiretskiy2013variational}
\begin{barticle}
\bauthor{\bsnm{Dragomiretskiy}, \binits{K.}},
\bauthor{\bsnm{Zosso}, \binits{D.}}:
\batitle{Variational mode decomposition}.
\bjtitle{IEEE transactions on signal processing}
\bvolume{62}(\bissue{3}),
\bfpage{531}--\blpage{544}
(\byear{2013})
\end{barticle}
\endbibitem

\bibitem[\protect\citeauthoryear{Bandara et~al.}{2021}]{bandara2021mstl}
\begin{botherref}
\oauthor{\bsnm{Bandara}, \binits{K.}},
\oauthor{\bsnm{Hyndman}, \binits{R.J.}},
\oauthor{\bsnm{Bergmeir}, \binits{C.}}:
Mstl: a seasonal-trend decomposition algorithm for time series with multiple
  seasonal patterns.
arXiv preprint arXiv:2107.13462
(2021)
\end{botherref}
\endbibitem

\bibitem[\protect\citeauthoryear{}{2023}]{engie}
\begin{botherref}
Engie Data Set
(2023).
\url{https://opendata-renewables.engie.com/explore/dataset/6eeb7f50-97f7-4ab2-8d36-c6d9f9491d84/information}
\end{botherref}
\endbibitem

\end{thebibliography}

\end{document}